\documentclass[manuscript,screen]{acmart}

\usepackage{tcolorbox}
\usepackage[utf8]{inputenc}
\usepackage{hyperref}
\usepackage{amsmath}
\usepackage{pifont}
\usepackage{wasysym}
\usepackage{todonotes}
\usepackage{comment}
\usepackage{pdflscape}
\usepackage{enumitem}
\usepackage{mdframed}
\usepackage[table]{xcolor}
\usepackage{subcaption}
\usepackage{fontawesome}
\usepackage{rotating}
\usepackage{adjustbox}
\usepackage{placeins}
\usepackage{float}
\usepackage{caption}
\usepackage{longtable}
\usepackage{booktabs}
\usepackage{multirow}
\usepackage{array}
\usepackage{tabularx}
\DeclareUnicodeCharacter{2212}{\textminus}
\usepackage{cleveref}
\usepackage{xcolor}

\usepackage{fancyhdr}

\usepackage{tikz}

\newcommand\copyrighttext{%
  \footnotesize © {Owner/Author} {2026}. This is the author's version of the work. It is posted here for your personal use. Not for redistribution. }
  
\newcommand\copyrightnotice{%
\begin{tikzpicture}[remember picture,overlay]
\node[anchor=south,yshift=10pt] at (current page.south) {\fbox{\parbox{\dimexpr\textwidth-\fboxsep-\fboxrule\relax}{\copyrighttext}}};
\end{tikzpicture}%
}

\newcolumntype{L}[1]{>{\raggedright\arraybackslash}p{#1}}

\AtBeginDocument{%
  }

\setcopyright{none}
\copyrightyear{}
\acmYear{}
\acmDOI{}

\acmJournal{TOSEM}
\renewcommand\footnotetextcopyrightpermission[1]{}

\makeatletter
\def\@ACM@acmPrice{}
\AtBeginDocument{%
  \def\ps@firstpagestyle{\ps@plain}%
  \def\ps@standardpagestyle{\ps@plain}%
  \pagestyle{plain}%
}
\makeatother

\begin{document}

\makeatletter
\def\ps@firstpagestyle{\ps@plain}
\def\ps@standardpagestyle{\ps@plain}
\makeatother
\title{Actionable Meta-Requirements for AI Robustness Auditing}

\author{Joshua Owotogbe}
\email{j.s.owotogbe@tilburguniversity.edu}
\orcid{0000-0001-9865-9310}
\affiliation{%
  \institution{Jheronimus Academy of Data Science}
  \city{'s-Hertogenbosch}
  \state{North Brabant}
  \country{Netherlands}
}
\affiliation{%
  \institution{Tilburg University}
  \city{Tilburg}
  \state{North Brabant}
  \country{Netherlands}
}

\author{Roya Nasiri}
\email{r.nasiri@tilburguniversity.edu}
\orcid{0009-0003-3985-3043}
\affiliation{%
  \institution{Jheronimus Academy of Data Science}
  \city{'s-Hertogenbosch}
  \state{North Brabant}
  \country{Netherlands}
}
\affiliation{%
  \institution{Tilburg University}
  \city{Tilburg}
  \state{North Brabant}
  \country{Netherlands}
}

\author{Indika Kumara}
\email{i.p.k.weerasinghadewage@tilburguniversity.edu}
\orcid{0000-0003-4355-0494}
\affiliation{%
  \institution{Jheronimus Academy of Data Science}
  \city{'s-Hertogenbosch}
  \state{North Brabant}
  \country{Netherlands}
}
\affiliation{%
  \institution{Tilburg University}
  \city{Tilburg}
  \state{North Brabant}
  \country{Netherlands}
}

\author{Damian Andrew Tamburri}
\email{datamburri@unisannio.it}
\orcid{0000-0003-1230-8961}
\affiliation{%
  \institution{Jheronimus Academy of Data Science}
  \city{'s-Hertogenbosch}
  \state{North Brabant}
  \country{Netherlands}
}
\affiliation{%
  \institution{University of Sannio}
  \city{Benevento}
  \state{Benevento}
  \country{Italy}
}

\author{Willem-Jan van den Heuvel}
\email{w.j.a.m.vdnHeuvel@tilburguniversity.edu}
\orcid{0000-0003-2929-413X}
\affiliation{%
  \institution{Jheronimus Academy of Data Science}
  \city{'s-Hertogenbosch}
  \state{North Brabant}
  \country{Netherlands}
}
\affiliation{%
  \institution{Tilburg University}
  \city{Tilburg}
  \state{North Brabant}
  \country{Netherlands}
}

\renewcommand{\shortauthors}{Owotogbe et al.}

\begin{abstract}
AI-enabled systems must remain dependable when the conditions under which they operate change (e.g., due to input perturbations, distribution shifts, or changes in external dependencies). This makes robustness auditing of such systems important for judging whether the available evidence is sufficient to support a defined robustness conclusion. However, designing an AI robustness-auditing framework is challenging. A key unresolved issue is determining the requirements such a framework should support; these requirements have not yet been consolidated. We address this gap through meta-requirements: higher-order requirements specifying what a class of auditing frameworks should support. In design science research, literature and domain experts are common sources of meta-requirements; accordingly, we derive the initial set from literature and validate and refine it with practitioners. Our systematic literature review retained 74 of 2{,}890 records and derived 43 meta-requirements across 16 categories. We evaluated them with 128 practitioners, obtaining 5{,}504 ratings and 586 substantive comments. Overall, 81.20\% of ratings indicated agreement or strong agreement. In particular, 41 of the 43 literature-derived meta-requirements were classified as strongly supported.  Practitioner feedback refined the requirements and identified an additional requirement concerning external components and supply-chain dependencies. The final set comprises 44 meta-requirements across 16 categories, of which 43 received
quantitative validation. These meta-requirements can help organizations implement AI robustness-auditing frameworks, assess existing frameworks for completeness, and refine them.
\end{abstract}






\maketitle

\copyrightnotice

\section{Introduction}
\label{sec:introduction}

AI-enabled systems are increasingly deployed in high-stakes domains such as healthcare, transportation, criminal justice, and financial services, where incorrect predictions, inconsistent decisions, or failures under unexpected operating conditions can affect safety, individual rights, and access to essential services \cite{rajpurkar2022ai,kuutti2020survey,dressel2018accuracy,lessmann2015benchmarking}. In this study, we use \emph{AI-enabled systems} to mean software systems in which at least one AI component contributes to the system's functionality \cite{martinez2022software}. Such systems may also depend on non-AI components, including data pipelines, services, frameworks, APIs, infrastructure, and other software elements that can influence their overall behaviour. For concision, we use \emph{AI auditing} throughout this paper to refer to the auditing of AI-enabled systems. In these domains, strong predictive performance under ordinary test conditions is insufficient. AI-enabled systems may produce different or degraded outputs when inputs are noisy or perturbed, data distributions shift, external dependencies change, or deployment conditions depart from those assumed during development \cite{rajpurkar2022ai,dressel2018accuracy,freiesleben2023beyond,tocchetti2025ai}. Robustness testing examines whether an AI-enabled system or its components maintain specified acceptable behaviour under defined changes to inputs or operating conditions. Such testing requires specifying the system or component being assessed, the expected behaviour, the disturbances or failure conditions considered, the evaluation procedure, and the acceptance criteria. \cite{freiesleben2023beyond,yang2020closer,hendrycks2019benchmarking,taori2020measuring,tocchetti2025ai}. Tests can produce observations about system behaviour under particular conditions, but an audit has a broader assurance role: it must judge whether the collected evidence is adequate for the stated robustness objective and make explicit the scope and limitations of that judgement.

This broader role makes the design of robustness auditing frameworks non-trivial. A useful framework must determine which concerns belong within the audit, what forms of evidence are relevant, how evaluation criteria should be established, how limitations should be documented, and when previously reached conclusions need to be revisited. These questions are increasingly consequential in regulatory and risk-management contexts. Article~15 of the European Union Artificial Intelligence Act (EU AI Act) requires high-risk AI systems to achieve an appropriate level of accuracy, robustness, and cybersecurity and to perform consistently in these respects \cite{EUAIAct2024,smuha2025regulation,nolte2025}.  Article~16 further requires providers to ensure compliance with these requirements; breaches of applicable provider obligations may be subject to administrative fines of up to EUR~15 million or, for undertakings, 3\% of their total worldwide
annual turnover for the preceding financial year, whichever is higher (Article~99(4)(a)) \cite{EUAIAct2024}. The National Institute of Standards and Technology Artificial Intelligence Risk Management Framework (NIST AI RMF), while voluntary, likewise addresses AI trustworthiness through characteristics including validity and reliability, safety, security, and resilience \cite{ai2023artificial}. Such instruments set important expectations, but they do not, by themselves, provide an operational specification for a robustness auditing framework.

Relevant knowledge for such a specification spans several research streams. Technical robustness research develops benchmarks, metrics, perturbation methods, attacks, defences, and evaluation procedures for examining AI-enabled systems under adversarial attacks, data corruptions, data distribution shifts, and other changes in relevant inputs or operating conditions
 \cite{hendrycks2019benchmarking,taori2020measuring,yang2020closer,li2023trustworthy,agarwal2025advancing,tocchetti2025ai}.
Research on AI auditing, in turn, examines audit scoping, evidence collection and interpretation, governance, and documentation \cite{raji2020closing,mokander2023auditing,minkkinen2022continuous,thelisson2024conformity}.  However, to the best of our knowledge, no study has systematically reviewed such literature from the relevant domains and derived a catalog of requirements that can guide the development and evaluation of robustness auditing frameworks for AI-enabled systems (Section~\ref{sec:rw_existing_limitations}, Table~\ref{tab:related_work_limitations}).  This study addresses this critical research gap. 

Design science research (DSR) offers a complementary perspective by expressing general design needs for an artifact, independently of a particular implementation
\cite{walls1992,zacharias2022designing,hevner2004}. Following the DSR terminology,  we use the term \emph{meta-requirements} to refer to these implementation-independent requirements that capture what a family of auditing frameworks for AI-enabled systems is expected to accommodate. In DSR, the literature provides a basis for identifying such requirements, while practitioner knowledge provides evidence to assess and refine their practical relevance.

To derive an actionable set of meta-requirements for an AI robustness auditing framework, inspired by DSR methods, we conducted a mixed-method study \cite{rabiser2010requirements,fahmideh2021software,waseem2021design} combining a systematic literature review with a practitioner validation survey. The database and supplementary searches identified 2{,}890 records, of which 74 met the eligibility criteria and quality-assessment threshold and were synthesised into 43 literature-derived meta-requirements across 16 categories and three dimensions. We then evaluated these requirements through a survey of 128 practitioners using Likert ratings and open-text feedback to examine support, refinement, challenge, and additional requirements emerging from professional experience. The findings show substantial convergence between the literature and practitioner perspectives. Overall, 81.20\% of the 5{,}504 ratings indicated agreement or strong agreement; 41 of the 43 literature-derived meta-requirements were classified as strongly supported, and the remaining two as supported. Practitioner feedback primarily refined wording, boundaries, access and evidence conditions, implementation assumptions, and lifecycle considerations, while also identifying an additional concern related to externally controlled dependencies. Integrating both evidence sources resulted in a final practitioner-informed set of 44 meta-requirements organised into 16 categories.

This article makes the following contributions:
\begin{itemize}[leftmargin=*]

    \item We derive a source-traceable set of AI robustness auditing meta-requirements from a systematic review of academic literature.

    \item We provide empirical evidence from practitioners on the perceived relevance and applicability of the literature-derived meta-requirements, using quantitative ratings and qualitative
    feedback to identify support, refinements, challenges, and previously underrepresented concerns.

    \item We discuss the implications of these findings for designing and evaluating future AI robustness auditing processes, methods, tools, and frameworks.

    \item We provide a replication package containing the review protocol, search strings, included studies, extraction forms, coding records, meta-requirement tables, survey instrument,
    and analysis outputs.

\end{itemize}

The remainder of this article is organised as follows. Section~\ref{sec:background} introduces the key concepts and scope of the study. Section~\ref{sec:related_work} reviews related work. Section~\ref{sec:method} describes the research methodology, including the systematic literature review, meta-requirement derivation, and practitioner survey. Sections~\ref{sec:rq1_results} and~\ref{sec:rq2_results} present the resulting meta-requirements and practitioner validation findings, respectively. Section~\ref{sec:discussion} discusses the findings and their implications. Section~\ref{sec:threats_validity} addresses threats to validity, and Section~\ref{sec:conclusion} concludes the article.

\section{Background}
\label{sec:background}

This section introduces the concepts needed to interpret the study. Table~\ref{tab:bg_concepts} summarises the terminology used throughout this article. The subsequent subsections show how robustness claims and evidence
inform auditing and meta-requirements.

\begin{table*}[!htbp]
\caption{Key concepts and definitions used in this study.}
\label{tab:bg_concepts}
\centering
\tiny
\renewcommand{\arraystretch}{1.08}
\begin{tabularx}{\textwidth}{p{3.2cm} X}
\toprule
\textbf{Concept} & \textbf{Meaning in this study} \\
\midrule

\textbf{AI-enabled system} &
A software system in which at least one AI component contributes to the system's functionality
\cite{martinez2022software}. \\

\textbf{AI audit} &
A systematic assessment of an AI-enabled system, its development or use, and relevant evidence against specified technical, organisational, legal, or ethical criteria
\cite{mokander2023auditing,minkkinen2022continuous}. \\

\textbf{AI robustness} &
The ability of an AI-enabled system to maintain specified acceptable behaviour when relevant inputs or operating conditions change
\cite{freiesleben2023beyond,tocchetti2025ai,wang2026survey}. \\

\textbf{Robustness testing} &
Testing whether an AI-enabled system or its components maintain specified acceptable behaviour under defined changes to inputs or operating conditions. \cite{wang2026survey,yu2026survey,haakansson2023towards,bayram2025towards,haakansson2023towards,silva2023case}\\

\textbf{Adversarial robustness} &
A form of robustness concerned with adversarial attacks, that is, deliberate attempts to cause incorrect or otherwise undesired system behaviour, for example by manipulating model inputs
\cite{wang2026survey,nolte2025,liu2022trustworthy}. \\

\textbf{Non-adversarial robustness} &
A form of robustness concerned with naturally occurring changes, such as noise, image blur, lighting changes, sensor degradation, or distribution
shift, where the data encountered during evaluation or deployment differ from those represented during model development
\cite{wang2026survey,nolte2025,freiesleben2023beyond,kemmerzell2025towards}. \\

\textbf{Threat model} &
The assumptions about what an attacker can know and what actions the attacker can perform against an AI-enabled system or model
\cite{chen2023ai}. \\

\textbf{Robustness metric} &
A quantitative measure used to report robustness, such as performance under specified test conditions, a robustness bound, or the probability
that a robustness condition is satisfied
\cite{wang2026survey}. \\

\textbf{Acceptance criterion} &
The condition used to determine whether an observed robustness result meets the required level of performance, such as a minimum accuracy, a
maximum permitted output change, or a maximum probability of error
\cite{wang2026survey}. \\

\textbf{Robustness claim} &
An explicit assertion that an identified system, model, or component will maintain stated acceptable behaviour under specified conditions and according to stated acceptance criteria
\cite{freiesleben2023beyond,simbeck2024they}. \\

\textbf{Robustness evidence} &
Information used to assess whether a robustness claim is supported, including test results, monitoring outputs, logs, analyses, and documentation
\cite{raji2020closing,momcilovic2026assuring,simbeck2024they}. \\

\textbf{AI robustness auditing} &
A structured assessment of whether available evidence supports a robustness claim within a defined assessment scope and against stated criteria
\cite{raji2020closing,casper2024black,simbeck2024they}. \\

\textbf{Meta-requirement for AI robustness auditing} &
A requirement that states what robustness auditing approaches should support across different AI-enabled systems, while allowing the specific method used to vary with the system and audit context  \cite{walls1992,hevner2004,zacharias2022designing,mantymaki2023designing}. \\

\textbf{Accuracy} &
The extent to which the outputs or predictions produced by an AI-enabled system are correct with respect to its intended task and the evaluation measure being used~\cite{nolte2025}. \\

\textbf{Computational cost} &
The amount of computational resources consumed or required to develop, evaluate, or operate an AI-enabled system, such as processing time, memory, energy, or computing capacity~\cite{chen2023ai}. \\

\textbf{Explainability} &
The extent to which the outputs, decisions, or behaviour of an AI-enabled system can be understood or explained to people who need to interpret, assess, or act on them~\cite{liu2022trustworthy}. \\

\textbf{Fairness} &
The extent to which an AI-enabled system avoids unjustified differences in outcomes for affected people or groups based on their characteristics
\cite{mehrabi1908survey,nanda2021fairness,liu2022trustworthy,wei2025trustworthy}. \\

\textbf{Bias} &
A systematic tendency in data, models, or system behaviour that can produce consistently different or poorer outcomes for particular people, groups, or cases, and may contribute to unfairness
\cite{mehrabi1908survey}. \\

\textbf{Safety} &
The extent to which the behaviour or outputs of an AI-enabled system avoid causing or contributing to harm to people, property, or the environment~\cite{liu2022trustworthy}. \\

\textbf{Privacy} &
The extent to which an AI-enabled system protects personal or sensitive information from unauthorized access, disclosure, leakage, or inference~\cite{wei2025trustworthy}. \\

\bottomrule
\end{tabularx}
\end{table*}

\subsection{AI Robustness and Robustness Claims}
\label{sec:Rob}

A robustness assessment must specify what is being assessed (e.g., a model or AI-enabled system), the behaviour or outcome to be evaluated together with its acceptance criterion (e.g., prediction accuracy and the minimum accuracy that must be maintained), and the condition under which it is evaluated
(e.g., adversarial inputs or distribution shift) \cite{freiesleben2023beyond,tocchetti2025ai,wang2026survey}. These conditions matter because robustness is condition-specific: evidence obtained under one type of change does not establish robustness under every other type of change
\cite{freiesleben2023beyond,nolte2025}. The robustness requirement being assessed must therefore be expressed as a specific claim before its supporting evidence can be evaluated. For example, an image-classification robustness claim may require a classifier to maintain a specified minimum accuracy when test images are
affected by Gaussian noise at a stated severity level; related tests may use blur, fog, or lighting changes to examine performance under other conditions
\cite{oala2020ml4h,wang2026survey}. Although this example targets a model, the assessment object may extend to the wider AI-enabled system. While the model represents the trained AI component, an AI-enabled system may also include services, frameworks, toolkits, platforms, infrastructure, and other software components \cite{yu2026survey,wang2026survey}. 
Accordingly, this study considers the robustness of the overall AI-enabled system, while allowing individual robustness claims to focus on a particular model or other system component.

\subsection{AI Robustness Auditing}
\label{sec:bg_auditing}

Once a robustness claim is specified, an AI robustness audit evaluates the available evidence against that claim, the assessment scope (i.e., the system or components included in the audit), and the criteria used to judge whether the claim is supported. The evidence may include test results, monitoring outputs, logs, data, documentation, and development or deployment records \cite{raji2020closing,casper2024black,simbeck2024they, momcilovic2026assuring}. Testing the robustness of an AI-enabled system or its components provides one type of audit evidence. For the image-classification claim above, the test could apply the specified noise to the test images and measure whether the classifier maintains the required accuracy \cite{wang2026survey}. Beyond such test results, the evidence available to the audit depends on the auditor's level of access. With black-box access, the auditor can query the system and observe its outputs; with white-box access, the auditor may also inspect internal model information, such as model parameters (weights) and intermediate computations (activations); and with outside-the-box access, development and deployment information, such as code, documentation, data, and internal evaluation results, may also be available \cite{casper2024black}.

\subsection{Meta-Requirements for Robustness Auditing}
\label{sec:bg_meta_requirements}

The preceding discussion shows why robustness auditing cannot rely on a single fixed procedure. The relevant claim, assessment object, available evidence, and level of auditor access can differ across AI-enabled systems, so a test or evaluation method suitable for one case may not be suitable for another \cite{freiesleben2023beyond,tocchetti2025ai,casper2024black,thelisson2024conformity}. Despite these differences, robustness audits share common needs,
including defining what is being assessed, obtaining and evaluating relevant evidence, and documenting the basis for conclusions
\cite{raji2020closing,casper2024black,simbeck2024they}. These differences in how robustness audits are conducted, together with the needs they share, motivate the use of meta-requirements to specify what robustness auditing approaches should address across different AI-enabled systems.

Design science research uses meta-requirements to express needs that should be addressed across a class of artefacts, methods, or approaches, while allowing different solutions to address those needs in different ways \cite{walls1992,hevner2004,gregor2007anatomy,lee2011theorizing, zacharias2022designing,mantymaki2023designing}. In this study, a robustness auditing meta-requirement is \emph{actionable} when it identifies a concern that can be examined in an audit and can guide the design or assessment of robustness auditing approaches, while the specific way in which that concern is addressed
may vary with the AI-enabled system and audit context. Practitioner evaluation assessed the perceived relevance and practical applicability of these meta-requirements, but their effectiveness when applied in real AI robustness audits has not yet been evaluated.

\section{Related Work}
\label{sec:related_work}

This section positions our study against prior work on challenges in robustness assessment, robustness practices across the AI lifecycle, and existing approaches to AI auditing.

\subsection{Challenges Relevant to Auditing AI Robustness}
\label{sec:rw_robustness_challenges}

Prior research highlights several challenges relevant to auditing AI robustness. First, robustness evaluations can differ in the attacks or input changes considered, their severity, the datasets used, the robustness metrics reported, and the acceptance criteria applied \cite{freiesleben2023beyond,wang2026survey,liu2022trustworthy}. These differences make results difficult to compare and can limit the extent to which evidence from one evaluation setting supports conclusions in another \cite{freiesleben2023beyond,javed2024robustness,kemmerzell2025towards}. Second, robustness failures can arise from interactions among multiple system layers and components rather than from the model alone \cite{yu2026survey,blood2023reliability}. This makes it difficult to determine whether model-level testing is sufficient, which components should be included in the audit scope, and what evidence is needed for a system-level conclusion \cite{raji2020closing,casper2024black,mokander2024auditing}. Third, thorough robustness assessment can require substantial computation, execution time, and infrastructure, particularly for verification, large-scale testing, and runtime monitoring \cite{wang2026survey,gupta2025monitoring}. These demands can limit the extent of testing and, consequently, the strength of the resulting audit evidence. Fourth, robustness interventions can involve trade-offs with predictive accuracy, efficiency, explainability, fairness, privacy, usability, or deployment cost \cite{yang2020closer,shafique2020robust,silva2020opportunities,nanda2021fairness,kaya2026explainable,tocchetti2025ai,chander2025toward}. Such trade-offs can complicate audit conclusions because improvements in robustness may affect other system properties that are also relevant to acceptance. Finally, robustness evidence can become outdated after deployment. Changes in production data, the deployed model, or deployment conditions can make earlier assessment results less representative of the current system and may require monitoring or reassessment \cite{bayram2025towards,li2025artificial,chen2023ai}. 
Together, these challenges motivate audit-oriented requirements addressing evaluation conditions, assessment scope, resource constraints, trade-offs, reassessment, and the basis and limitations of robustness conclusions.

\subsection{Robustness Practices Across the AI Lifecycle}
\label{sec:rw_robustness_practices}

Prior research places robustness-related activities across the AI lifecycle, including data preparation, model training, robustness evaluation, deployment, and continuous monitoring \cite{li2023trustworthy,bayram2025towards,tang2026advancing}. First, at the data-preparation stage, relevant practices include checking and cleaning missing, inconsistent, or noisy data; augmenting the available data with additional examples; and constructing test data that represent relevant conditions, such as noise, blur, fog, lighting changes, or differences between development and deployment data \cite{silva2023case,oala2020ml4h,wang2026survey,javed2024robustness,tang2026advancing}. Second, during model training, robustness can be improved through adversarial training, which exposes the model to adversarial examples; robust optimisation, which minimises worst-case loss under bounded input changes; regularisation techniques such as dropout; and architecture modifications designed to improve robustness \cite{yang2020closer,xiong2022towards,agarwal2025advancing,mohseni2022taxonomy,javed2024robustness,tang2026advancing}. Third, robustness evaluation examines how a trained model behaves under specified conditions. It can include adversarial testing, testing with common input changes, and evaluation on data from different distributions \cite{wang2026survey,javed2024robustness}. Formal or statistical verification can additionally assess whether a specified robustness condition holds within a defined input region or with a stated probability \cite{wang2026survey}. These assessment activities examine whether a trained model satisfies specified robustness conditions and should therefore be distinguished from techniques intended to improve robustness \cite{freiesleben2023beyond}. Finally, after deployment, robustness-related activities include estimating prediction confidence or uncertainty, rejecting predictions when confidence is insufficient, detecting inputs that differ from the training distribution or have been deliberately manipulated, and monitoring model behaviour during operation \cite{mohseni2022taxonomy,gupta2025monitoring}. Production processes can also monitor distribution or concept drift, retain relevant documentation and logs, and define when model updates or retraining are required \cite{bayram2025towards,li2025artificial}.  Across the AI-enabled system lifecycle, robustness auditing must relate evidence to the assessed robustness claim, system or component within scope, acceptance criteria, and assessment conditions \cite{raji2020closing,mokander2024auditing,thelisson2024conformity}. It must also document evidence limitations and determine when changes require reassessment. These needs motivate the consolidation of technical and procedural knowledge into audit-oriented meta-requirements.

\subsection{Limitations of Prior Work on AI Robustness Auditing}
\label{sec:rw_existing_limitations}

Prior work relevant to AI robustness auditing contributes complementary foundations from different research streams. Robustness-focused studies clarify how robustness can be understood and evaluated. Tocchetti et al.~\cite{tocchetti2025ai} systematise robustness terminology and organise prior work around robustness-improvement, application-specific, and evaluation approaches, while Freiesleben and Grote~\cite{freiesleben2023beyond} develop a conceptual account of robustness and distinguish robustness testing within an audit from techniques intended to improve robustness. Together, these studies provide a basis for robustness assessment, but they do not define the requirements of a complete robustness auditing process. Studies on AI auditing provide guidance on how audits are organised and carried out. Raji et al.~\cite{raji2020closing} propose a five-stage internal audit process that moves from defining the audit scope and examining the system context to collecting evidence, testing the system, and reflecting on the findings. Mökander et al.~\cite{mokander2024auditing} organise LLM auditing across the organisation developing the model, the model itself, and its downstream applications. Simbeck~\cite{simbeck2024they} structures auditing from defining the scope and assessment criteria through evaluation to subsequent monitoring. Together, these studies show how AI audits can be organised, but they do not define robustness-specific audit requirements.

Regulatory and trustworthy-AI studies identify requirements and obligations that AI-enabled systems are expected to satisfy. Nolte et al.~\cite{nolte2025} analyse the robustness and cybersecurity requirements of the EU AI Act and their implementation implications. Thelisson and Verma~\cite{thelisson2024conformity} examine how high-risk AI systems can be assessed against regulatory requirements before market entry and propose supporting assessment tools. Díaz-Rodríguez et al.~\cite{diaz2023connecting} analyse seven requirements for trustworthy AI, including technical robustness and safety. Together, these studies clarify what AI-enabled systems should satisfy, but not the meta-requirements needed to design and conduct robustness audits. Requirements-oriented studies provide the closest methodological precedent. M\"antym\"aki et al.~\cite{mantymaki2023designing} derive seven meta-requirements for organisational AI governance through design science research, combining literature-based premises with insights from an industry-academia project involving
organisational design partners. Zacharias et al.~\cite{zacharias2022designing} derive three meta-requirements for explainable feature selection and evaluate a resulting software artefact with practitioners and researchers. Von Zahn et al.~\cite{von2025navigating} derive four meta-requirements for assessing AI conformity, with their implemented assessment centred on fairness, explainability, and performance. Boufaied et al.~\cite{boufaied2025practitioner} provide practitioner evidence on how fairness requirements are interpreted and managed across the AI/ML lifecycle.

These studies provide useful methods for deriving and evaluating requirements, but the reviewed work does not consolidate robustness-specific audit requirements for stating what robustness is being claimed, which parts of the AI-enabled system are covered, which failures or operating changes should be tested, what evidence is needed to support the audit conclusion, and when reassessment is required. We also did not identify a study that derives such meta-requirements from the literature and evaluates them with practitioners (Table~\ref{tab:related_work_limitations}).

\begin{table*}[htbp]
\centering
\caption{Comparison of representative studies relevant to AI robustness auditing meta-requirements.}
\label{tab:related_work_limitations}

\tiny
\setlength{\tabcolsep}{3pt}
\renewcommand{\arraystretch}{1.15}

\resizebox{\textwidth}{!}{%
\begin{tabular}{
p{0.075\textwidth}
c
p{0.20\textwidth}
p{0.15\textwidth}
c
c
p{0.13\textwidth}
p{0.085\textwidth}
p{0.34\textwidth}}
\hline

\textbf{Ref.} &
\textbf{Year} &
\textbf{Objective} &
\textbf{Method} &
\textbf{Robustness Focus} &
\textbf{Audit/ Assur.} &
\textbf{Req./MR Output} &
\textbf{Pract. Eval.} &
\textbf{Findings / Limitation} \\
\hline

\cite{raji2020closing} &
2020 &
Develop an end-to-end internal framework for algorithmic auditing for development of AI systems. &
Framework synthesis informed by auditing practices in adjacent disciplines; illustrative cases &
\LEFTcircle &
\ding{51} &
- &
\ding{55} &
Provides a broad internal AI auditing framework with staged evidence collection and testing, but no robustness-specific meta-requirements. \\
\hline

\cite{zacharias2022designing} &
2022 &
Develop a design framework for explainable feature selection and instantiate it as a software artefact. &
Design Science Research; literature and legal-document review; two expert focus groups &
\ding{55} &
\ding{55} &
3 meta-requirements + design principles &
\ding{51} ($n=12$) &
Derives and evaluates meta-requirements for explainable feature selection, not robustness auditing. \\
\hline

\cite{mantymaki2023designing} &
2023 &
Derive meta-requirements for organisational AI governance frameworks. &
Design Science Research; literature-derived premises; industry-academia collaboration with organisational design partners &
\ding{55} &
\LEFTcircle &
7 meta-requirements &
\ding{51} &
Derives meta-requirements for AI governance with organisational design-partner input, but does
not develop robustness-specific auditing meta-requirements. \\
\hline

\cite{freiesleben2023beyond} &
2023 &
Develop a unifying conceptual account of ML robustness based on targets and modifiers. &
Conceptual analysis &
\ding{51} &
\LEFTcircle &
- &
\ding{55} &
Provides a conceptual robustness framework and distinguishes robustness testing from robustness improvement, but no robustness auditing requirements framework. \\
\hline

\cite{diaz2023connecting} &
2023 &
Connect trustworthy-AI principles, ethics, regulation, and technical requirements. &
Multidisciplinary conceptual analysis &
\LEFTcircle &
\LEFTcircle &
7 trustworthy-AI requirements analysed &
\ding{55} &
Analyses seven trustworthy-AI requirements, including robustness and safety, but does not derive robustness auditing meta-requirements. \\
\hline

\cite{mokander2024auditing} &
2023 &
Develop a three-layered blueprint for auditing large language models. &
Literature review; conceptual synthesis &
\LEFTcircle &
\ding{51} &
7 design claims; no meta-requirements &
\ding{55} &
Proposes coordinated governance, model, and application audits and seven design claims, but no robustness-specific meta-requirements. \\
\hline

\cite{simbeck2024they} &
2024 &
Develop an auditing framework for learning analytics systems covering fairness, transparency, and robustness. &
Conceptual synthesis of audit approaches and domain-specific criteria &
\ding{51} &
\ding{51} &
- &
\ding{55} &
Provides a four-phase audit process in which robustness is assessed alongside fairness and transparency, but no robustness-specific meta-requirements. \\
\hline

\cite{tocchetti2025ai} &
2025 &
Reconcile terminology concerning AI robustness and organise robustness-improvement and evaluation approaches from a human-centred perspective. &
Systematic survey; thematic analysis &
\ding{51} &
\ding{55} &
- &
\ding{55} &
Systematises robustness concepts, methods, applications, and evaluation approaches, but does not develop an framework of audit requirements. \\
\hline

\cite{nolte2025} &
2025 &
Analyse robustness and cybersecurity requirements in the EU AI Act and their implementation implications. &
Legal analysis of AI Act provisions; comparison with ML literature &
\ding{51} &
\LEFTcircle &
Legal requirements analysed &
\ding{55} &
Clarifies legal robustness and cybersecurity requirements and implementation challenges, but does not develop robustness auditing meta-requirements. \\
\hline

\cite{von2025navigating} &
2025 &
Develop and evaluate a design framework and software artefact for AI conformity assessment. &
Design Science Research; literature and legal-document review; seven expert interviews; five focus groups &
\ding{55} &
\ding{51} &
4 meta-requirements + design principles &
\ding{51} ($n=21$) &
Provides four practitioner-evaluated meta-requirements for AI conformity assessment, with the implemented assessment centred on fairness and explainability rather than robustness. \\
\hline

\cite{boufaied2025practitioner} &
2026 &
Investigate how practitioners interpret and manage fairness requirements across the AI/ML software-development lifecycle. &
Qualitative interview study; 26 semi-structured interviews; thematic analysis &
\ding{55} &
\ding{55} &
Fairness-requirement findings + challenges framework &
\ding{55} &
Provides practitioner evidence on fairness requirements, but does not derive or evaluate robustness auditing meta-requirements. \\
\hline

\textbf{This study} &
\textbf{2026} &
\textbf{Derive literature-grounded meta-requirements for AI robustness auditing and evaluate them with practitioners.} &
\textbf{Systematic literature review + practitioner survey} &
\textbf{\ding{51}} &
\textbf{\ding{51}} &
\textbf{43 literature-derived + 1 practitioner-informed meta-requirement} &
\textbf{\ding{51} ($n=128$)} &
\textbf{Derives 43 literature-based meta-requirements, evaluates them with practitioners, and incorporates one additional meta-requirement from qualitative feedback, resulting in a final set of 44 meta-requirements.} \\
\hline

\end{tabular}%
}

\vspace{0.5em}

\begin{minipage}{\textwidth}
\footnotesize
\textit{Note.}
\ding{51} = explicitly addressed;
\LEFTcircle = partially addressed;
\ding{55} = not addressed; - = no requirements or meta-requirements reported.
\textit{Robustness Focus} indicates whether robustness is a central concern of the study.
\textit{Audit/Assur.} indicates explicit audit, assurance, or regulatory-assessment coverage.
\textit{Req./MR Output} distinguishes analysed requirements from explicitly derived meta-requirements and design principles.
\textit{Pract. Eval.} indicates empirical evaluation of a proposed framework, requirements, or artefact with practitioners or domain experts.
\end{minipage}

\end{table*}

\section{Methodology}
\label{sec:method}

\begin{figure*}[t]
    \centering
    \includegraphics[
        width=\textwidth,
        keepaspectratio
    ]{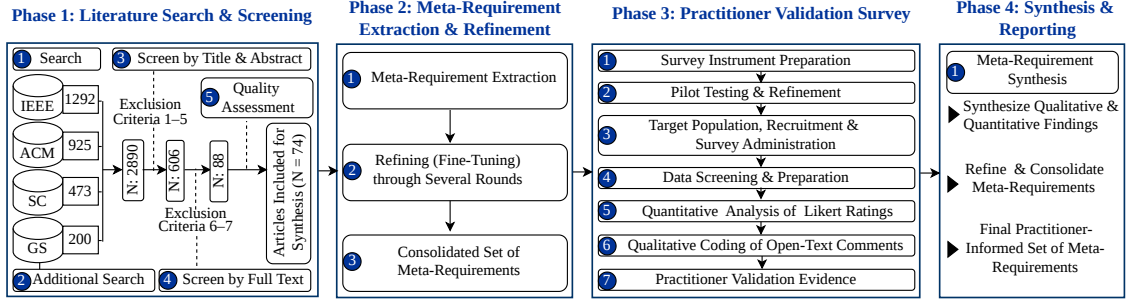}
    \caption{Overview of the four-phase research method used to derive, refine, validate, and synthesize the meta-requirements.}
    \label{fig:mrq_method}
\end{figure*}

This study adopts a mixed-methods design \cite{rabiser2010requirements,fahmideh2021software,waseem2021design} to derive and validate meta-requirements for AI robustness auditing. It combines a systematic literature review, used to derive the literature-grounded meta-requirements, with a practitioner validation survey, used to assess and refine them based on professional experience. The review component was guided by established systematic review guidelines \cite{kitchenham2007guidelines,garousi2017experience,garousi2017software}. The research process comprises four phases: (1) literature search and screening, (2) meta-requirement extraction and refinement, (3) practitioner validation survey, and (4) meta-requirement synthesis and reporting (see Figure~\ref{fig:mrq_method}).

\subsection{Research Questions}
\label{sec:research_questions}

This study is guided by two research questions. RQ1 identifies the meta-requirements for AI robustness auditing, while RQ2 examines practitioners' perspectives on those meta-requirements (see Table~\ref{tab:research_questions}).

\begin{table}[htbp]
  \caption{Research questions and motivation.}
  \label{tab:research_questions}
  \tiny
  \begin{tabular}{|p{0.39\textwidth}|p{0.56\textwidth}|}
    \hline
    \textbf{Research Question} & \textbf{Motivation} \\
    \hline

    \textbf{RQ1:} What meta-requirements should an AI robustness auditing framework satisfy? & This question identifies the requirements that should guide the design and conduct of AI robustness auditing. \\
     \hline

    \textbf{RQ2:} What are practitioners' perspectives on the proposed meta-requirements for AI robustness auditing? & This question examines practitioners' support for, qualifications of, challenges to, and suggested refinements or additions to the proposed meta-requirements. \\


    \hline
  \end{tabular}
\end{table}

RQ1 is addressed through Phases~1 and~2, which identify, code, and consolidate audit-relevant evidence from the literature into meta-requirements. RQ2 is addressed through Phases~3 and~4, which analyse practitioner ratings and open-text comments and integrate this validation evidence into the final set of meta-requirements.

\subsection{Phase 1: Literature Search and Screening}
\label{sec:literature_search}

Following established systematic-review guidance~\cite{kitchenham2007guidelines}, we constructed the search around AI/ML systems, robustness-related concerns, and audit-relevant assessment activities. To focus on studies with practical relevance to robustness auditing, the search strings additionally required terms referring to practitioners (practitioner, developer, engineer, or stakeholder) and terms referring to empirical studies (interview or survey). Pilot searches indicated that removing these restrictions substantially increased the number of irrelevant records. We therefore retained the restrictions to improve search precision, while recognising that they could exclude relevant studies that did not use this terminology. Candidate studies were retrieved from three academic databases: IEEE Xplore ($N = 1{,}292$), ACM Digital Library ($N = 925$), and Scopus ($N = 473$) (see Phase~1 in Figure~\ref{fig:mrq_method}). We selected these databases because they provide broad coverage of software engineering, artificial intelligence, and information systems research~\cite{chen2010towards}. The search was restricted to studies published between January 2019 and February 2026, and all results were exported in BibTeX and CSV formats for reference management and data processing. Table~\ref{tab:search_results} reports the final search strings and per-database retrieval counts.

\begin{table}[!htbp]
  \caption{Search Strings and Results}
  \label{tab:search_results}
  \tiny
  \centering
  \renewcommand{\arraystretch}{1.15}
  \begin{tabularx}{\textwidth}{p{1.4cm} X p{1.7cm}}
    \hline
    \textbf{Digital Library} 
    & \textbf{Search String} 
    & \textbf{Retrieved Articles Numbers} \\
    \hline

    ACM
    &
    [[Abstract: "robustness"] OR [Abstract: resilien*] OR [Abstract: reliab*] OR [Abstract: dependab*]]
    AND [[Abstract: "artificial intelligence"] OR [Abstract: "machine learning"] OR [Abstract: "algorithmic"] OR [Abstract: "ai"] OR [Abstract: "ml"]]
    AND [[Abstract: audit*] OR [Abstract: assess*] OR [Abstract: test*] OR [Abstract: evaluat*]]
    AND [[Abstract: practitioner*] OR [Abstract: developer*] OR [Abstract: engineer*] OR [Abstract: stakeholder*]]
    AND [[Abstract: interview*] OR [Abstract: survey*]]
    & 925 \\
    \hline

    Scopus
    &
    TITLE-ABS-KEY (("robustness" OR "resilien*" OR "reliab*" OR "dependab*")
    AND ("artificial intelligence" OR "machine learning" OR "algorithmic" OR "AI" OR "ML")
    AND ("audit*" OR "assess*" OR "test*" OR "evaluat*")
    AND ("practitioner*" OR "developer*" OR "engineer*" OR "stakeholder*")
    AND ("interview*" OR "survey*"))
    & 473 \\
    \hline

    IEEE Xplore
    &
    ("All Metadata":"robustness" OR "All Metadata":resilien* OR "All Metadata":reliab* OR "All Metadata":dependab*)
    AND ("All Metadata":"artificial intelligence" OR "All Metadata":"machine learning" OR "All Metadata":"algorithmic" OR "All Metadata":"AI" OR "All Metadata":"ML")
    AND ("All Metadata":audit* OR "All Metadata":assess* OR "All Metadata":test* OR "All Metadata":evaluat*)
    AND ("All Metadata":practitioner* OR "All Metadata":developer* OR "All Metadata":engineer* OR "All Metadata":stakeholder*)
    AND ("All Metadata":interview* OR "All Metadata":survey*)
    & 1292 \\
    \hline

    \textbf{Total}
    & -
    & \textbf{2690} \\
    \hline
  \end{tabularx}
\end{table}

To complement the database searches, we conducted a supplementary Google Scholar search, following prior systematic reviews that used it to extend database-based retrieval~\cite{gaber2024malware,ajaykumar2021survey,ihle2023incentive}. We examined the first 20 pages using terms reflecting the same AI/ML, robustness, audit, practitioner, and empirical-study scope as the database searches, and applied the same eligibility criteria. The database searches yielded 2,690 records, and the supplementary Google Scholar search added 200, resulting in 2,890 records for title/abstract screening. Following established SLR guidance~\cite{islam2019multi,butijn2020blockchains,garousi2019guidelines}, 
Table~\ref{tab:criteria} presents the inclusion and exclusion criteria.

\begin{table}[htbp]
  \caption{Inclusion and exclusion criteria.}
  \label{tab:criteria}
  \tiny
  \begin{tabular}{|p{0.47\textwidth}|p{0.48\textwidth}|}
    \hline
    \textbf{Inclusion Criteria} & \textbf{Exclusion Criteria} \\
    \hline

    \textbf{IC1:} Studies must be written in English. &
    \textbf{EC1:} Studies not written in English. \\
    \hline

    \textbf{IC2:} Studies must be published between January 2019 and February 2026. &
    \textbf{EC2:} Studies published outside the selected time window. \\
    \hline

    \textbf{IC3:} Studies must be available in full text. &
    \textbf{EC3:} Studies whose full text is unavailable or inaccessible. \\
    \hline

    \textbf{IC4:} Studies must be academic research publications, including journal articles, conference papers, or workshop papers of at least six pages. &
    \textbf{EC4:} Duplicate records, papers shorter than six pages, books, editorials, abstracts, posters, proceedings summaries, presentation-only papers, or other publication types lacking sufficient research content. \\
    \hline
    
    \textbf{IC5:} Studies must examine AI, machine learning, deep learning, large language models, algorithmic systems, or AI-enabled software systems as the object of study. &
    \textbf{EC5:} Studies outside the AI/ML domain, or studies that use AI/ML only as an analytical tool for prediction, classification, diagnosis, forecasting, optimisation, sensing, or domain analysis without examining the AI/ML system itself. \\
    \hline
    
    \textbf{IC6:} Studies must address robustness or related dependability concerns of AI/ML systems, such as reliability, resilience, safety, fault tolerance, failure behaviour, distribution shift, out-of-distribution behaviour, adversarial robustness, or failure analysis. &
    \textbf{EC6:} Studies that use robustness, reliability, or resilience only in a non-AI sense, or only to describe general predictive performance, efficiency, usability, adoption, trust, sustainability, educational outcomes, physical-system performance, or organisational resilience. \\
    \hline
    
    \textbf{IC7:} Studies must connect AI/ML robustness or dependability to audit-relevant activities, including assessment, testing, evaluation, verification, validation, monitoring, assurance, compliance, governance, documentation, or evidence generation. &
    \textbf{EC7:} Studies focused only on adjacent topics such as fairness, privacy, explainability, cybersecurity, traceability, forensics, ethics, policy, adoption, education, usability, or general trust without a direct connection to AI/ML robustness or dependability assessment. \\
    \hline

  \end{tabular}
\end{table}

The selection process proceeded in five steps (see Phase~1 in Figure~\ref{fig:mrq_method}). Two researchers independently screened all 2,890 records at title/abstract level using IC1–IC5 and EC1–EC5, achieving 96.23\% agreement (Cohen’s $\kappa=0.882$). The 109 disagreements were resolved by consensus, leaving 606 studies for full-text review. The same researchers independently assessed the 606 full texts using IC6–IC7 and EC6–EC7, achieving 94.22\% agreement ($\kappa=0.741$); we resolved 35 disagreements by consensus. After reconciliation, 88 studies were retained for quality assessment.

\subsubsection{Quality Assessment}
\label{sec:quality_assessment}

The 88 eligible studies were assessed using six quality criteria adapted from prior SLR practices \cite{bajaj2019systematic,browning2024systematic}. The criteria examined whether: (QA1) the study objective or research question was clearly stated; (QA2) the study design or methodology was suitable for that objective; (QA3) the AI/ML system context was sufficiently described; (QA4) the data or evidence sources and analysis procedures were sufficiently described; (QA5) the findings and conclusions were supported by the reported evidence and analysis; and (QA6) relevant assumptions, limitations, or threats to validity were reported. Each criterion was scored as \emph{Yes} (1), \emph{Partly} (0.5), or \emph{No} (0). Studies scoring at least 4 out of 6 were retained for the final synthesis. The first researcher scored all 88 studies against QA1-QA6, and the second researcher reviewed the completed quality assessments for all studies. Any differences identified during this review were discussed and resolved by consensus. After quality assessment, 74 studies were retained.
The QA1-QA6 scores, total quality score, and final inclusion decision for each of the 88 assessed studies are provided in the replication package (see the Data Availability statement).

\begin{figure}[htbp]
    \centering

    \begin{minipage}[t]{0.50\textwidth}
        \centering
        \includegraphics[
            width=\linewidth,
            keepaspectratio
        ]{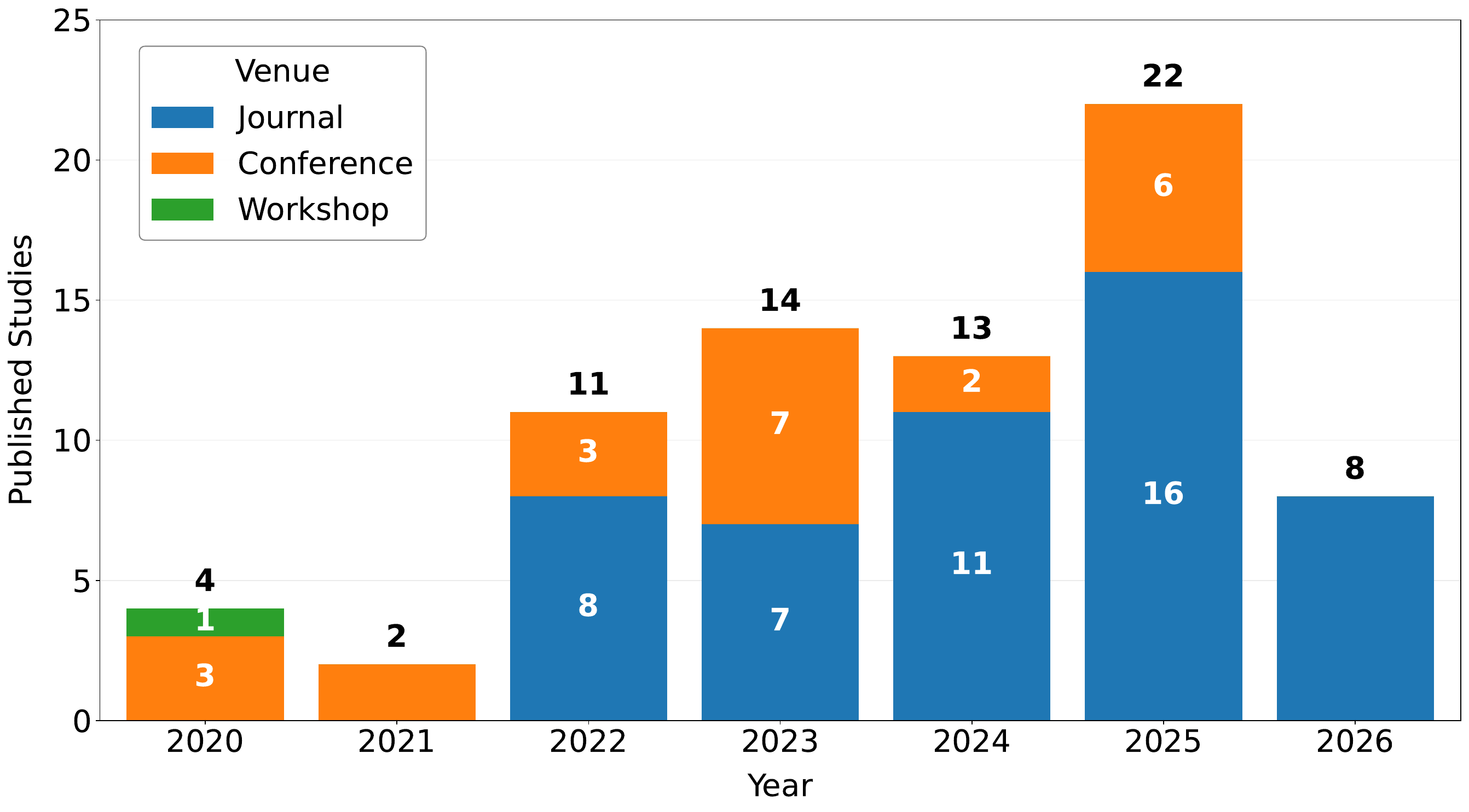}
        \caption{Annual distribution of the reviewed studies by publication venue type.}
        \label{fig:venues_by_year}
    \end{minipage}
    \hfill
    \begin{minipage}[t]{0.44\textwidth}
        \centering
        \includegraphics[
            width=\linewidth,
            keepaspectratio
        ]{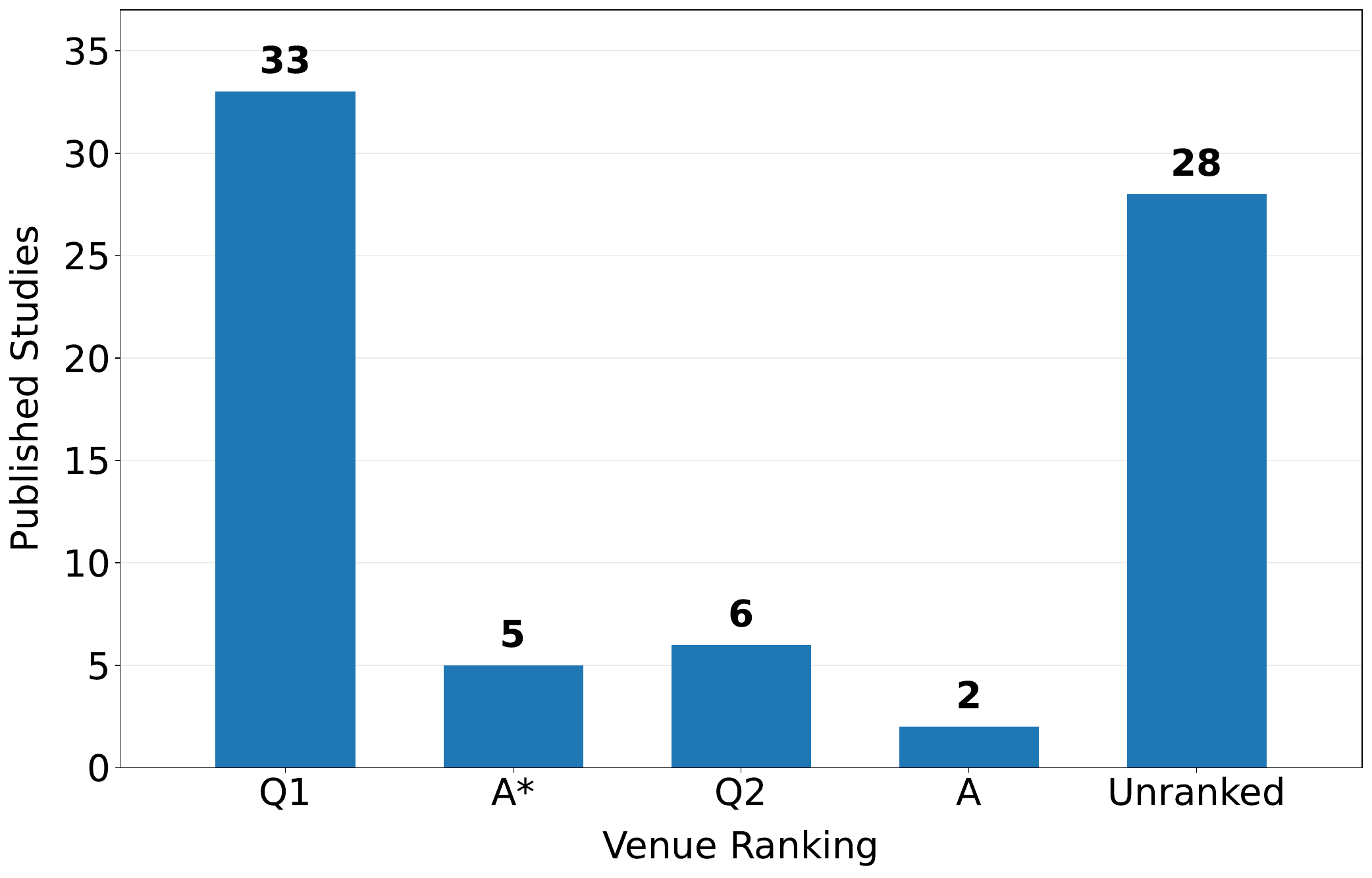}
        \caption{Distribution of the reviewed studies by venue ranking.}
        \label{fig:venue_ranking}
    \end{minipage}

\end{figure}

Among the 74 studies retained after quality assessment, publication activity was limited in 2020 ($N=4$) and 2021 ($N=2$), followed by 11 studies in 2022, 14 in 2023, 13 in 2024, and a peak of 22 in 2025 (Figure~\ref{fig:venues_by_year}). The 2026 count ($N=8$) reflects the February 2026 search cut-off. Journal articles accounted for 50 studies, followed by conference papers ($N=23$) and one workshop paper. Figure~\ref{fig:venue_ranking} shows that 33 studies appeared in Q1 venues and 28 in unranked venues, followed by Q2 ($N=6$), A* ($N=5$), and A ($N=2$) venues. Venue ranking was reported descriptively and was not used as a proxy for study quality or as an exclusion criterion. Review rigour was instead supported through predefined eligibility criteria, independent screening, consensus resolution, and traceable evidence extraction. Following Galster et al.~\cite{galster2013variability}, lower-ranked and unranked studies were retained when they met the eligibility criteria and contributed relevant evidence.



\subsection{Phase 2: Meta-Requirement Extraction and Refinement}
\label{sec:mr_extraction}

The 74 retained studies were analysed using a structured qualitative coding procedure to derive meta-requirements for AI robustness auditing \cite{nekrasov2025iac,charmaz2006constructing}. The coding focused on how studies define robustness claims, specify audit scope, evaluate robustness, document evidence, assign audit responsibilities, involve stakeholders, monitor deployed systems, report limitations, and support reassessment. For each study, we used a standard extraction form to record the bibliographic details, AI/ML context, robustness concern, audit or evaluation activity, supporting quotation, extracted pattern, candidate category, and associated meta-requirement. The completed extraction records are provided in the replication package (see the Data Availability statement) and link each supporting quotation to its extracted pattern, candidate category, and literature-derived meta-requirement.

\begin{figure}[!ht]
    \centering
    \includegraphics[
        width=0.85\columnwidth,
        keepaspectratio
    ]{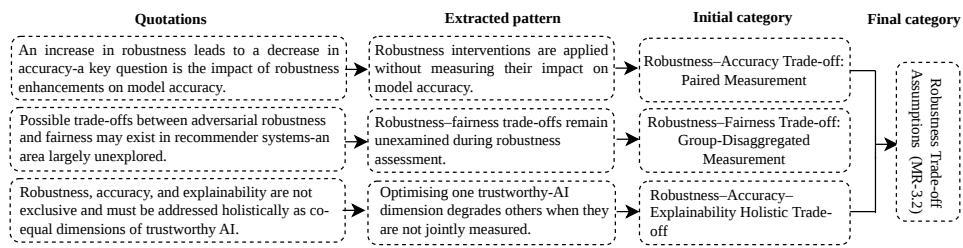}
    \caption{Example of how extracted literature evidence is coded and consolidated into MR-3.2, Robustness Trade-off Assumptions.}
    \label{fig:mr_coding_example}
\end{figure}

The first stage applied open coding \cite{charmaz2006constructing} to identify audit-relevant passages and summarise each as an extracted pattern. In the second stage, axial coding grouped related patterns by underlying audit concern, including patterns that spanned different domains and system contexts. For example, evidence concerning the effects of robustness interventions on accuracy, fairness, and explainability was progressively grouped into the broader concern of trade-off measurement (see Figure~\ref{fig:mr_coding_example}). 

The third stage converted the candidate categories into meta-requirements using explicit merge, separation, and retention criteria, following prior studies that consolidated lower-level evidence into higher-level tasks, categories, or requirements \cite{fahmideh2021software,nekrasov2025iac,rabiser2010requirements}. Candidate categories were merged when they represented the same auditing requirement despite differences in terminology, technique, or application context. They were kept separate when merging them would remove a different audit responsibility, evidence requirement, or audit decision. A candidate category was retained as a meta-requirement when it was within the study scope, supported by traceable source evidence, applicable beyond a single technique or system, and not already represented by another meta-requirement.

The resulting categories were assigned to three higher-level dimensions according to their primary purpose, following the use of explicit classification criteria in prior empirical studies \cite{sanchez2022mutation,ustunboyacioglu2024data}. Categories were classified as \emph{Human} when they primarily concerned people, roles, governance, competence; as \emph{Functional} when they primarily concerned audit activities or capabilities; and as \emph{Quality} when they primarily concerned properties of the auditing approach itself. Where a category related to more than one dimension, its dominant purpose determined the assignment. Recurrence across studies increased the supporting evidence for a concern but was not used as a fixed inclusion threshold. A concern could therefore be retained when it met the criteria above even if it appeared in relatively few studies, while repeated evidence for the same higher-level concern did not result in additional meta-requirements.

Following established principles for writing quality requirements \cite{heck2010software,heck2014quality}, we refined the resulting meta-requirements for clarity, lack of ambiguity, atomicity, unique identification, linkage to supporting source evidence, audit assessability, and consistency across the set. We formulated each meta-requirement as an abstract, technology-agnostic statement specifying what a robustness auditing framework should support, require, or make explicit, and assigned each an identifier, category, and requirement statement. One researcher conducted the extraction, coding, and initial meta-requirement formulation. A second researcher independently reviewed the extracted patterns, category assignments, and resulting meta-requirements. The team resolved disagreements and ambiguities through discussion and consensus. This process produced 43 literature-derived meta-requirements across 16 categories, which formed the basis of the practitioner validation survey in Phase~3. Table~\ref{tab:all_mr_final_refined} reports these together with MR-8.4, which was added after the survey and was therefore not rated by respondents. The replication package contains the corresponding extraction records, including the source passage, extracted pattern, category assignment, and resulting meta-requirement, allowing readers to inspect how the source evidence was consolidated.

\subsection{Phase 3: Practitioner Validation Survey}
\label{sec:practitioner_validation}

The 43 meta-requirements derived in Phase~2 were evaluated through a practitioner survey addressing RQ2. Following established survey guidance in software engineering, we combined Likert-scale ratings with optional open-text feedback to assess practitioner support, interpretation, and suggested refinements, challenges, and additional concerns~\cite{kitchenham2002designing,kitchenham2002constructing,linaker2015guidelines}. Comparable studies have similarly used practitioner surveys to assess and refine previously identified requirements and framework elements~\cite{van2019practitioners,zellmer2026decision}.

\subsubsection{Survey Instrument Preparation} 
\label{sec:survey_design} 
We developed a Google Forms questionnaire comprising 44 numbered questions. First, the survey opened with a study-information and consent section describing the research purpose, expected completion time, voluntary participation, confidentiality arrangements, and researcher contact information. Second, Q2-Q5 collected demographic information, including age, gender, country or region, and education, while Q6-Q11 captured participants' professional profile, including AI/ML experience, robustness-assessment or auditing experience, academic and professional experience levels, AI/ML-related role, and involvement in robustness-related activities.

Third, Q12-Q44 formed the validation section. The 43 meta-requirements were distributed across 16 matrix questions, one for each validation category, with each meta-requirement presented as a separate rating row. Respondents rated each meta-requirement using a five-point Likert scale from 1~=~\emph{strongly disagree} to 5~=~\emph{strongly agree}. Each matrix question was followed by an optional open-text question for comments or suggested changes, and the
final question invited feedback on missing requirements, duplication, unclear wording, implementation barriers, or other improvements. The complete questionnaire is provided in the replication package (see the Data Availability statement).

\subsubsection{Pilot Testing and Refinement}
\label{sec:pilot_testing}

Following prior practitioner-survey practice \cite{van2019practitioners,waseem2021design}, we piloted the survey from 24 June to 5 July 2026 with eight participants with relevant AI/ML, robustness, or auditing experience. The pilot assessed completion time, clarity of instructions and terminology, interpretation of the category descriptions and meta-requirements, suitability of the five-point scale, possible overlap, and usability across devices. Reported completion times ranged from approximately 15 to 30 minutes, with some reviewers finding completion more difficult on mobile devices. Based on the feedback, we shortened the introductory text, clarified category descriptions and meta-requirement labels, simplified wording, and retained the five-point scale and optional definitions. The eight pilot participants were not included in the main survey sample,
and their responses were excluded from the main dataset and analysis.

\subsubsection{Target Population, Recruitment, and Survey Administration}
\label{sec:participant_recruitment}

Eligible participants were researchers and practitioners worldwide with academic, professional, or applied experience in AI/ML systems and related activities, including robustness assessment, model evaluation, adversarial evaluation, AI auditing, AI safety, model reliability, or assurance.  We used purposive non-probability sampling because the study required participants with relevant experience in areas related to AI robustness auditing, while the specialised target population could not be comprehensively enumerated for probability sampling \cite{kitchenham2008personal,fahmideh2021software,waseem2021design,canedo2023privacy}. Potential participants were identified through GitHub and LinkedIn, channels that have also been used to recruit specialised practitioners in prior software engineering surveys \cite{fahmideh2021software,waseem2021design,van2019practitioners}. On GitHub, we searched relevant repositories and contributors using robustness-related queries and GitHub's \emph{Best match} ranking. Representative queries included \texttt{"AI robustness"}, \texttt{"robustness auditing"}, \texttt{"adversarial robustness"}, and \texttt{"ML robustness"}. Approximately 1,200 potential participants were identified through GitHub and 520 through LinkedIn. Survey invitations were sent to all 1,720 identified candidates. We received 143 responses, corresponding to a response rate of 8.31\%. After applying the predefined response-screening criteria, 15 responses were excluded, leaving 128 valid responses for analysis. Survey invitations briefly introduced the study, stated the eligibility criteria, and directed potential participants to the study-information and consent page of the questionnaire (see Section~\ref{sec:survey_design}). The study received ethical clearance from the Research Ethics and Data Management Committee of the Tilburg School of Humanities and Digital Sciences, Tilburg University (REDC 2026.04). The main survey was administered through Google Forms from 8 July to 15 August 2026, and
informed consent was obtained before participation. After completing the survey, respondents could optionally enter a draw for a \euro100 gift card \cite{de2022software}. Contact information submitted for the draw was collected separately from the survey responses.

\subsubsection{Data Screening and Preparation}
\label{sec:data_preparation}

Following prior practitioner-survey research \cite{waseem2021design}, we defined inclusion and exclusion criteria for screening the survey responses, as shown in Table~\ref{tab:survey_screening}. The inclusion criterion required respondents to demonstrate relevant academic, professional, or practical experience in at least one of the study's target areas, as assessed from their professional-profile responses. Responses containing inconsistent, apparently random, or meaningless answers were considered for exclusion, as were responses without informed consent. We received 143 responses to the main survey. After applying the predefined screening criteria, 15 responses were excluded under E1, leaving 128 valid responses for quantitative analysis. The responses were exported from Google Forms and processed in Python by identifying the 43 meta-requirement rating columns, standardising response labels, converting the five-point Likert responses to values from 1 to 5, checking missing values, assigning anonymous respondent identifiers, and transforming the ratings into long format. The resulting dataset contained 5{,}504 ratings, with no missing meta-requirement values. Open-text responses from the 128 valid respondents were screened separately because completion of these fields was optional. Across the 16 category-level comment fields and the final overall-comment field, 670 entries were extracted. Empty responses, uninterpretable statements, and comments without analytically usable feedback were removed, leaving 586 substantive comments for qualitative coding and synthesis. Each retained comment was linked to its anonymous respondent identifier, source category, and position within the corresponding response batch to preserve traceability.

\begin{table}[htbp]
\centering
\caption{Inclusion and exclusion criteria for selecting valid responses.}
\label{tab:survey_screening}
\tiny
\begin{tabularx}{\linewidth}{p{0.12\linewidth} p{0.08\linewidth} X}
\toprule

\textbf{Inclusion} &
\textbf{I1:} &
The respondent must demonstrate relevant academic, professional, or practical experience in AI/ML systems, AI robustness assessment or auditing, robustness testing, model evaluation, adversarial evaluation, AI safety, model reliability, or related assurance activities. \\

\midrule

\textbf{Exclusion} &
\textbf{E1:} &
A response is excluded if it contains inconsistent, apparently random, or meaningless answers. \\

&
\textbf{E2:} &
A response without informed consent is excluded.\\

\bottomrule
\end{tabularx}
\end{table}

\subsubsection{Quantitative Analysis of Likert Ratings}
\label{sec:quantitative_analysis}

The 5{,}504 Likert ratings were analysed descriptively at three levels: the overall response distribution, the individual meta-requirement level, and the category level. This type of item-level and aggregated analysis has been used in prior practitioner surveys \cite{vogelsang2020views}. At the overall level, we calculated the frequencies and percentages of all five response options. At the meta-requirement level, each of the 43 literature-derived meta-requirements was analysed using the 128 corresponding participant ratings. At the category level, ratings were aggregated across the meta-requirements belonging to each of the 16 survey categories; these percentages summarise all ratings within a category and are not respondent-level agreement rates.

For each meta-requirement and category, we reported response frequencies and percentages for agreement, neutral, and disagreement, as well as the median and interquartile range (IQR). Because individual Likert items are ordinal, the median and IQR were used as the primary descriptive summaries \cite{agresti2010analysis,van2019practitioners}. Mean, standard deviation (SD), and mode were reported only as supplementary descriptive measures. Consistent with prior practitioner-survey studies, the quantitative ratings were interpreted alongside open-text feedback to assess support for the proposed requirements and identify refinements or missing concerns \cite{fahmideh2021software,van2019practitioners}.

Ratings of 4 (\emph{agree}) and 5 (\emph{strongly agree}) were grouped as agreement-level responses, a rating of 3 as neutral, and ratings of 1 (\emph{strongly disagree}) and 2 (\emph{disagree}) as disagreement-level responses. To provide a consistent descriptive interpretation of practitioner support, we defined study-specific quantitative-status rules for each meta-requirement. Prior consensus studies have used percentage-agreement thresholds to distinguish levels of agreement, including 60\% as a minimum consensus threshold and 75\% as a commonly used consensus threshold \cite{olivero2022delphi,diamond2014defining}. A meta-requirement was classified as \emph{strongly supported} when its median was at least 4 and at least 75\% of ratings were agreement-level responses, and as \emph{supported} when its median was at least 4 and agreement was at least 60\% but below 75\%. A meta-requirement was classified as \emph{mixed or unclear} when its median was 3 or at least 40\% of ratings were neutral, and as \emph{challenged} when its median was at most 3 and at least 30\% of ratings were disagreement-level responses. Cases not meeting these criteria were designated as requiring qualitative interpretation. These thresholds were analytic decision rules defined for this study, not inferential significance criteria. The quantitative results were interpreted together with the open-text comments to determine the extent to which the literature-derived meta-requirements were supported, refined, challenged, or extended \cite{fahmideh2021software,van2019practitioners}. No inferential comparisons between demographic subgroups were performed because the survey was designed to evaluate the meta-requirements rather than test differences between respondent groups.

{

\tiny
\begin{longtable}{L{0.8cm}L{1.9cm}L{7.8cm}L{3.0cm}}
\caption{Meta-requirements for AI robustness auditing: 43 literature-derived requirements and MR-8.4, added after the practitioner survey.}
\label{tab:all_mr_final_refined} \\
\toprule
\textbf{ID} & \textbf{Sub-Category} & \textbf{Requirement} & \textbf{Source(s)} \\
\midrule
\endfirsthead
\multicolumn{4}{l}{\small\textit{(continued from previous page)}} \\
\toprule
\textbf{ID} & \textbf{Sub-Category} & \textbf{Requirement} & \textbf{Source(s)} \\
\midrule
\endhead
\midrule
\multicolumn{4}{r}{\small\textit{(continued on next page)}} \\
\endfoot
\bottomrule
\endlastfoot

\multicolumn{4}{l}{\textbf{Human Dimension}} \\
\midrule
\multicolumn{4}{l}{\textbf{Category 1: Standards and Governance}} \\
\midrule

MR-1.1 & Standardized Metrics &
Robustness auditing should use standardised robustness metrics and benchmark definitions where available, so that reported robustness results are comparable across systems, audits, and regulatory contexts. &
\cite{nolte2025, khinvasara2024robustness, oala2020ml4h, blood2023reliability, yu2026survey, tocchetti2025ai, yang2020closer, agarwal2025advancing, chen2023ai, madaio2022assessing, freiesleben2023beyond, casper2024black, raji2020closing, thelisson2024conformity, nanda2021fairness, hornischer2026robustness, javed2024robustness, liu2022trustworthy, li2023trustworthy, wei2025trustworthy, costanza2022audits} \\
\midrule

MR-1.2 & Standardized Lexicon and Terminology &
Robustness audits should use a shared vocabulary, taxonomy, and terminology for robustness and auditability concepts. &
\cite{mohseni2022taxonomy, vadlapati2026ai, yu2026survey, ge2024survey, tocchetti2025ai, yang2020closer, chen2023ai, freiesleben2023beyond, raji2020closing, thelisson2024conformity, nanda2021fairness, nolte2025, khinvasara2024robustness, hornischer2026robustness, liu2022trustworthy, li2023trustworthy, wei2025trustworthy, costanza2022audits} \\
\midrule

MR-1.3 & Standardized Tools &
Robustness audits should use common or comparable tools, checklists, templates, scorecards, libraries, or toolkits where they support consistent assessment. &
\cite{raji2020closing, momcilovic2026assuring, ojewale2025towards, oala2020ml4h, tocchetti2025ai, dev2025building, yu2026survey, ge2024survey, yang2020closer, agarwal2025advancing, casper2024black, kokina2025challenges, thelisson2024conformity, nolte2025, khinvasara2024robustness, javed2024robustness, li2023trustworthy, costanza2022audits} \\
\midrule

MR-1.4 & Actionable Guidance &
Practitioners should receive practical guidance for selecting tests, thresholds, evidence, mitigations, and audit steps. &
\cite{kokina2025challenges, khinvasara2024robustness, narula2025exploring, blood2023reliability, habbal2024artificial, thelisson2024conformity, cooper2022accountability, laux2024trustworthy, chen2023ai, saeed2024security, xiong2022towards, mccormack2025trust, koshiyama2024towards, momcilovic2026assuring, chander2025toward} \\
\midrule

MR-1.5 & Robustness-aware Integration Guidance &
Robustness guidance should support embedding robustness into development, deployment, monitoring, and assurance workflows. &
\cite{li2023trustworthy, yu2026survey, tocchetti2025ai, chen2023ai, madaio2022assessing, freiesleben2023beyond, casper2024black, kokina2025challenges, raji2020closing, thelisson2024conformity, oala2020ml4h, nolte2025, khinvasara2024robustness, hornischer2026robustness, javed2024robustness, liu2022trustworthy, wei2025trustworthy, costanza2022audits} \\
\midrule

MR-1.6 & Governance Protocols &
Organisational or regulatory protocols should define responsibility, escalation, remediation, accountability, and risk decisions for robustness assurance. &
\cite{habbal2024artificial, liu2022trustworthy, li2025artificial, cooper2022accountability, raji2020closing, costanza2022audits, mokander2024auditing, laux2024trustworthy, mccormack2025trust, chamola2023review,  thelisson2024conformity, nolte2025} \\
\midrule

MR-1.7 & Audit Access Policy &
Audit access policy should specify the artefacts and access level required for robustness assessment, including relevant datasets, documentation, source code, model or system information, deployment details, logs, internal evaluation results, APIs, and conditions for black-box, white-box, or outside-the-box access. &
\cite{li2025artificial, costanza2022audits, raji2020closing, koshiyama2024towards, simbeck2024they, thelisson2024conformity, casper2024black, ojewale2025towards, muller2025practical, chander2025toward, gupta2025monitoring} \\
\midrule

MR-1.8 & Auditor Protocols &
Auditor protocols should specify how the audit is conducted, including auditor conduct, independence, method transparency, evidence handling, reporting format, and conditions under which audit results may be interpreted. &
\cite{raji2020closing, costanza2022audits, ojewale2025towards, casper2024black, thelisson2024conformity, oala2020ml4h, mokander2024auditing, ge2024survey, tocchetti2025ai, chen2023ai, madaio2022assessing, freiesleben2023beyond, kokina2025challenges, nanda2021fairness, nolte2025, khinvasara2024robustness, liu2022trustworthy, li2023trustworthy} \\
\midrule

\multicolumn{4}{l}{\textbf{Category 2: Engagement}} \\
\midrule

MR-2.1 & Diverse Stakeholders &
Robustness audits should identify and involve relevant technical, domain, governance, user, and affected-party perspectives, including users or groups exposed to robustness failures, according to the robustness risks and impacts covered by the audit scope. &
\cite{tocchetti2025ai, muller2025practical, chander2025toward, gupta2025monitoring, habbal2024artificial, costanza2022audits, li2025artificial, balendran2025scoping, mokander2024auditing, zahid2025explainability, raji2020closing, silva2023case, laux2024trustworthy, momcilovic2026assuring, ojewale2025towards, madaio2022assessing, lee2023adversarial, mccormack2025trust, robertson2023angler, dev2025building, hancox2020robustness, nanda2021fairness, nolte2025, hornischer2026robustness, liu2022trustworthy,kaur2022trustworthy} \\
\midrule

MR-2.2 & Stakeholder Involvement Prioritization &
Stakeholder involvement should be prioritised according to each group's exposure to robustness failures, domain knowledge, decision authority, and relevance to deployment risks. &
\cite{agarwal2025advancing, madaio2022assessing, kokina2025challenges, raji2020closing, thelisson2024conformity, oala2020ml4h, nolte2025, li2023trustworthy, costanza2022audits} \\
\midrule

MR-2.3 & Independent Auditor &
Robustness audits should include independent review with conflict control, methodological transparency, and sufficient access for the intended assurance claim. &
\cite{ojewale2025towards, casper2024black, li2025artificial, liu2022trustworthy, thelisson2024conformity, costanza2022audits, mccormack2025trust, kokina2025challenges, raji2020closing, habbal2024artificial, laux2024trustworthy, mokander2024auditing} \\
\midrule

MR-2.4 & Early-stage Involvement &
Domain and technical experts should be involved during problem framing, design, and assessment planning when early expert judgement is needed to define robustness-relevant assumptions, risks, or evidence. &
\cite{raji2020closing, li2025artificial, mccormack2025trust, cooper2022accountability, madaio2022assessing, hornischer2026robustness,kaur2022trustworthy} \\
\midrule

\multicolumn{4}{l}{\textbf{Category 3: Robustness Definition}} \\
\midrule

MR-3.1 & Scoping Robustness Concerns &
The robustness scope should define the audit claim, system boundary, disturbance types, failure modes, deployment contexts, assumptions, and justified exclusions. &
\cite{dev2025building, xiong2022towards, chander2025toward, ge2024survey, narula2025exploring, balendran2025scoping, hornischer2026robustness, liu2022trustworthy,  blood2023reliability, gupta2025monitoring, mohseni2022taxonomy, saqib2024comprehensive, tocchetti2025ai, pan2023ai, saeed2024security, javed2024robustness, habbal2024artificial, mccormack2025trust, freiesleben2023beyond, wei2025trustworthy, vadlapati2026ai, li2023trustworthy, casper2024black, wang2026survey, palazzolo2024measurable, kemmerzell2025towards, zahid2025explainability, tsingenopoulos2025adaptive, chen2023ai, kaya2026explainable, silva2023case, ouyang2021ai, chamola2023review, bayram2025towards, petersen2022responsible, haakansson2023towards, roy2024enhancing, tang2026advancing, raji2020closing, koshiyama2024towards, mokander2024auditing, nolte2025,  khinvasara2024robustness, lee2023adversarial, yang2020closer,  costanza2022audits, taesiri2022visual, simbeck2024they,  robertson2023angler, oala2020ml4h, chen2025ai, agarwal2025advancing} \\
\midrule

MR-3.2 & Robustness Trade-off Assumptions &
The audit should identify assumptions about how the scoped robustness claim may affect, depend on, or conflict with other system properties, including accuracy, fairness, privacy, explainability, safety, cost, and affected-user outcomes, and should document how these assumptions shape measurement, interpretation, and decision-making. &
\cite{li2025artificial, hardy2025more, liu2022trustworthy, nolte2025, chander2025toward, balendran2025scoping, nanda2021fairness, yang2020closer,  simbeck2024they, koshiyama2024towards, javed2024robustness, chen2025ai, ge2024survey, bayram2025towards, tocchetti2025ai, li2023trustworthy, habbal2024artificial, vadlapati2026ai, tsingenopoulos2025adaptive, saeed2024security, palazzolo2024measurable, mohseni2022taxonomy, kemmerzell2025towards, freiesleben2023beyond, thelisson2024conformity, haakansson2023towards, dev2025building, oala2020ml4h, mccormack2025trust, chen2023ai, costanza2022audits, pan2023ai, lefcourt2022robustness, lee2023adversarial, madaio2022assessing, ouyang2021ai,  kaya2026explainable, tang2026advancing, mokander2024auditing,kaur2022trustworthy} \\
\midrule

\multicolumn{4}{l}{\textbf{Category 4: Knowledge and Skills}} \\
\midrule

MR-4.1 & Multidisciplinary Skills &
Robustness audits should involve combined AI, machine-learning, software, security, data, domain, social, legal, risk, and audit expertise where required by the assessment. &
\cite{costanza2022audits, li2025artificial, vadlapati2026ai, ala2024trusting, muller2025practical, kokina2025challenges, laux2024trustworthy, robertson2023angler, tocchetti2025ai, madaio2022assessing, mccormack2025trust, nolte2025, mokander2024auditing, khinvasara2024robustness, xiong2022towards, li2023trustworthy, silva2023case,kaur2022trustworthy} \\
\midrule

MR-4.2 & Robustness-awareness Educational Programmes and Training &
Personnel should receive education or training on robustness risks, methods, monitoring, mitigation, communication, and tool use relevant to their roles. &
\cite{zahid2025explainability, muller2025practical, chander2025toward,  koshiyama2024towards,  ala2024trusting, habbal2024artificial, geleta2023maestro, kokina2025challenges} \\
\midrule

\multicolumn{4}{l}{\textbf{Functional Dimension}} \\
\midrule
\multicolumn{4}{l}{\textbf{Category 5: Data Auditability}} \\
\midrule

MR-5.1 & Data Coverage and Sufficiency &
Robustness assessment data should sufficiently cover relevant populations, data types, modalities, subgroups, scenarios, domains, edge cases, and operating conditions. &
\cite{javed2024robustness, tocchetti2025ai, yang2020closer, agarwal2025advancing, chen2023ai, madaio2022assessing, yu2026survey, freiesleben2023beyond, kokina2025challenges, raji2020closing, nanda2021fairness, oala2020ml4h, nolte2025, hornischer2026robustness, li2023trustworthy, wei2025trustworthy, costanza2022audits} \\
\midrule

MR-5.2 & Data Description, Provenance, and Lineage &
Robustness audits should require dataset-level descriptions and provenance records that identify data sources, acquisition conditions, processing steps, annotation basis, known limitations, and links between the data and the robustness claim.  &
\cite{javed2024robustness, lefcourt2022robustness, bayram2025towards, ge2024survey, casper2024black, kokina2025challenges, raji2020closing, thelisson2024conformity, oala2020ml4h, nolte2025, khinvasara2024robustness, li2023trustworthy, wei2025trustworthy, costanza2022audits} \\
\midrule

MR-5.3 & Data Quality &
Data used for robustness assessment should be checked for noise, missingness, errors, corruption, leakage, imbalance, representativeness, and other data defects. &
\cite{ouyang2021ai, blood2023reliability, lefcourt2022robustness, bayram2025towards, narula2025exploring, oala2020ml4h, mccormack2025trust, martinez2022software, yang2020closer, hardy2025more, silva2023case, mohseni2022taxonomy} \\
\midrule

MR-5.4 & Data Subgroup and Bias Source Identification &
Data subgroups, target attributes, representation gaps, and data-level bias sources should be identified where they may affect the credibility of robustness findings across relevant populations, contexts, or operating conditions. &
\cite{roy2024enhancing,  li2023trustworthy, oala2020ml4h, kokina2025challenges, mccormack2025trust, kemmerzell2025towards, khinvasara2024robustness, alzubaidi2023towards, mokander2024auditing,kaur2022trustworthy} \\
\midrule

MR-5.5 & Robustness Pre-processing Mitigation &
The audit should assess whether data-level mitigation
methods, such as cleaning, filtering, transformation,
augmentation, or recovery of contaminated training
data, address identified robustness risks and improve
robustness under the assessed conditions. &
\cite{blood2023reliability, dev2025building, palazzolo2024measurable, tocchetti2025ai, chen2023ai, freiesleben2023beyond, javed2024robustness, li2023trustworthy, wei2025trustworthy} \\
\midrule

\multicolumn{4}{l}{\textbf{Category 6: Model Auditability}} \\
\midrule

MR-6.1 & Model Robustness Probing &
Auditors should probe model behaviour with defined robustness stressors, including perturbations, adversarial examples, stress cases, distribution shifts, benchmark scenarios, and complementary test methods where one probing method is insufficient. &
\cite{ouyang2021ai, kou2024beyond,  martinez2022software, mokander2024auditing, oala2020ml4h, yang2020closer, robertson2023angler, vadlapati2026ai, lee2023adversarial, nolte2025, hardy2025more, kaya2026explainable, petersen2022responsible, agarwal2025advancing, wang2026survey, tocchetti2025ai, costanza2022audits, silva2023case, tsingenopoulos2025adaptive, saqib2024comprehensive, nanda2021fairness, dev2025building, ge2024survey, hancox2020robustness} \\
\midrule

MR-6.2 & Model Intended Use &
The model's intended application, users, operating conditions, automation level, limitations, and unsuitable uses should be specified for the audited model. &
\cite{madaio2022assessing, raji2020closing, thelisson2024conformity, oala2020ml4h, nolte2025} \\
\midrule

MR-6.3 & Model Explainability &
Model behaviour, decision logic, and robustness failures should be interpretable or otherwise explainable enough to support the audit. &
\cite{ouyang2021ai, javed2024robustness, methnani2024s, hancox2020robustness, kaya2026explainable, tocchetti2025ai, khinvasara2024robustness, habbal2024artificial, casper2024black, tang2026advancing, saqib2024comprehensive, mccormack2025trust, zahid2025explainability, kemmerzell2025towards, liu2022trustworthy, chander2025toward, saeed2024security, chen2025ai, petersen2022responsible, mohseni2022taxonomy, roy2024enhancing,  oala2020ml4h, taesiri2022visual, ala2024trusting, li2023trustworthy, ge2024survey,  chamola2023review, kokina2025challenges, koshiyama2024towards, pan2023ai, cooper2022accountability, li2025artificial,kaur2022trustworthy} \\
\midrule

MR-6.4 & Model Bias and Failure Source Identification &
Model-level causes of robustness failure, including shortcut learning, overfitting, spurious correlations, and unstable features, should be identified where relevant. &
\cite{lee2023adversarial, petersen2022responsible, robertson2023angler, hancox2020robustness, freiesleben2023beyond, casper2024black, kemmerzell2025towards} \\
\midrule

MR-6.5 & Robustness In-processing Mitigation &
The audit should assess whether training-time mitigation methods, such as adversarial training, robust optimisation, regularisation, or adapted loss functions, address identified robustness risks and improve robustness under the
assessed conditions. &
\cite{yang2020closer, agarwal2025advancing, javed2024robustness, wei2025trustworthy, tocchetti2025ai} \\
\midrule

\midrule

\multicolumn{4}{l}{\textbf{Category 7: Output Auditability}} \\
\midrule

MR-7.1 & Robustness Post-processing Mitigation&

Robustness audits should identify output-level unstable, unsafe, biased, harmful, or inconsistent behaviour under robustness stress, trace the likely source of the failure, and assess whether post-processing mitigations reduce the identified robustness risk before or during deployment. &

\cite{yu2026survey, chen2023ai, javed2024robustness, liu2022trustworthy, tocchetti2025ai} \\
\midrule

\multicolumn{4}{l}{\textbf{Category 8: Workflow Integration}} \\
\midrule

MR-8.1 & Software Development Process Integration &
Robustness checks should be integrated into requirements, design, implementation, testing, deployment, maintenance, and review activities. &
\cite{silva2023case, mohseni2022taxonomy, chen2023ai, li2023trustworthy, laux2024trustworthy, martinez2022software, alzubaidi2023towards, tang2026advancing, blood2023reliability,kaur2022trustworthy} \\
\midrule

MR-8.2 & MLOps Frameworks Integration &
Robustness tests, alerts, retraining triggers, monitoring, and reassessment steps should be embedded into ML or MLOps pipelines where applicable. &
\cite{koshiyama2024towards, narula2025exploring, dev2025building, wei2025trustworthy, bayram2025towards, ge2024survey} \\
\midrule

MR-8.3 & Robustness-aware Integration Compatibility &
Robustness tools and procedures should be compatible with relevant software, data, model, deployment, monitoring, and incident-management infrastructure. &
\cite{koshiyama2024towards, tocchetti2025ai, martinez2022software, dev2025building, kokina2025challenges, thelisson2024conformity, khinvasara2024robustness} \\
\midrule

MR-8.4$^{*}$ & External Component and Supply Chain Assurance &
Robustness audits should identify external models, data, software, services, and infrastructure on which the audited AI-enabled system depends; assess how their failure or change could affect system behaviour; and state how limited access to those dependencies restricts the robustness conclusion. & Practitioner feedback (P079, P102); supported by prior studies \cite{dev2025building,yu2026survey,martinez2022software,blood2023reliability,casper2024black,mokander2024auditing} \\
\midrule
\multicolumn{4}{l}{\textbf{Category 9: Human-in-the-Loop}} \\
\midrule

MR-9.1 & Robustness-aware Human Participation Levels &
Robustness audits should define human participation levels for robustness-relevant decisions, including review, validation, intervention, override, or approval, and should state how available access to inputs, outputs, procedures, transparency information, model internals, or other evidence limits the resulting robustness judgement. &
\cite{tocchetti2025ai, kokina2025challenges, chamola2023review, mccormack2025trust, kou2024beyond, oala2020ml4h, habbal2024artificial, chen2023ai, casper2024black, thelisson2024conformity,kaur2022trustworthy} \\
\midrule

\multicolumn{4}{l}{\textbf{Category 10: Evaluation}} \\
\midrule

MR-10.1 & Continuous Robustness Assessment &
Robustness should be formally reassessed repeatedly or when triggered by changes in data, users, attacks, models, or deployment conditions. &
\cite{thelisson2024conformity, blood2023reliability, momcilovic2026assuring, laux2024trustworthy, chen2023ai, raji2020closing, li2025artificial, mokander2024auditing, simbeck2024they, khinvasara2024robustness, mccormack2025trust, costanza2022audits, wang2026survey, madaio2022assessing, koshiyama2024towards} \\
\midrule

MR-10.2 & Robustness-aware Progress Tracking and Prioritisation &
Robustness evaluations should maintain a prioritised record of completed checks, unresolved robustness concerns, follow-up tests, mitigation needs, and reassessment actions, so that remaining audit work is visible and sequenced. &
\cite{costanza2022audits, mokander2024auditing, dev2025building, tocchetti2025ai, chen2023ai} \\
\midrule

MR-10.3 & Cross-Layer Failure Validation &
Selected failures should be validated across relevant data, model, software, service, infrastructure, and human-interaction layers. &
\cite{raji2020closing, blood2023reliability, mohseni2022taxonomy, yu2026survey, dev2025building} \\
\midrule

\multicolumn{4}{l}{\textbf{Category 11: Monitoring}} \\
\midrule

MR-11.1 & Continuous, Automated, and Real-time Robustness Monitoring &

Deployed AI-enabled systems should maintain a defined post-deployment robustness monitoring process that tracks robustness indicators, applies predefined thresholds, issues automated or real-time alerts for specified robustness violations, and triggers reassessment when system behaviour, data, deployment conditions, or threat conditions change. &

\cite{gupta2025monitoring, blood2023reliability, nolte2025, habbal2024artificial, chen2023ai, bayram2025towards, li2025artificial, mohseni2022taxonomy, thelisson2024conformity, robertson2023angler, koshiyama2024towards, khinvasara2024robustness, narula2025exploring, vadlapati2026ai, momcilovic2026assuring, mccormack2025trust, martinez2022software, chen2025ai, roy2024enhancing, schmelczer2023trustworthy, raji2020closing, ala2024trusting, methnani2024s, chamola2023review, haakansson2023towards,  kaya2026explainable, ojewale2025towards} \\
\midrule

MR-11.2 & Robustness-aware Logging &
Inputs, outputs, model behaviour, errors, alerts, updates, and context should be logged where needed to diagnose robustness failures. &
\cite{thelisson2024conformity, schmelczer2023trustworthy, methnani2024s, raji2020closing, li2025artificial} \\
\midrule

\multicolumn{4}{l}{\textbf{Category 12: Communication}} \\
\midrule

MR-12.1 & Robustness Dashboards and Visualisation &
Robustness dashboards and visualisations should present audit-relevant metrics, model behaviour, monitoring trends, subgroup outcomes, failures, alerts, and supporting evidence in a form that helps intended users inspect and interpret robustness evidence. &
\cite{xiong2022towards, robertson2023angler, tocchetti2025ai, methnani2024s, ala2024trusting, li2025artificial, muller2025practical, nolte2025, agarwal2025advancing, kokina2025challenges, raji2020closing, oala2020ml4h} \\
\midrule

MR-12.2 & Robustness Documentation and Risk Report &
Robustness audits should document and report the audit objective, evidence base, evaluated system and datasets, robustness findings, limitations, mitigations, residual risks, reproducibility conditions, traceability information, and audience-appropriate explanations needed to interpret the audit conclusion. &
\cite{koshiyama2024towards, mokander2024auditing, laux2024trustworthy, costanza2022audits, ojewale2025towards, mccormack2025trust, raji2020closing, momcilovic2026assuring, li2023trustworthy, chander2025toward, li2025artificial, vadlapati2026ai, chen2023ai, petersen2022responsible, oala2020ml4h, kemmerzell2025towards, palazzolo2024measurable, tsingenopoulos2025adaptive, ouyang2021ai,  alzubaidi2023towards, bayram2025towards,  yu2026survey, cooper2022accountability, simbeck2024they, gupta2025monitoring, blood2023reliability, zahid2025explainability, kaya2026explainable, methnani2024s, hancox2020robustness, freiesleben2023beyond,  taesiri2022visual, casper2024black, dev2025building, ala2024trusting} \\
\midrule

\multicolumn{4}{l}{\textbf{Quality Dimension}} \\
\midrule
\multicolumn{4}{l}{\textbf{Category 13: Efficiency}} \\
\midrule

MR-13.1 & Resource Efficiency and Scalability &
A robustness auditing process should be cost-effective, time-efficient, and scalable, so that organisations with different resource capacities can assess, monitor, mitigate, and report robustness risks within practical audit-cycle constraints and across the models, datasets, perturbations, users, and deployment settings covered by the audit scope. &
\cite{kou2024beyond, kokina2025challenges, tocchetti2025ai, madaio2022assessing, nanda2021fairness, khinvasara2024robustness, javed2024robustness, wei2025trustworthy, kaya2026explainable, saqib2024comprehensive, thelisson2024conformity, martinez2022software, gupta2025monitoring, mccormack2025trust, chen2025ai, mohseni2022taxonomy, wang2026survey, mokander2024auditing, hornischer2026robustness, dev2025building} \\
\midrule

\multicolumn{4}{l}{\textbf{Category 14: Adaptability}} \\
\midrule

MR-14.1 & Domain-specific Adaptability &
Robustness definitions, perturbations, thresholds, tests, and interpretations should be tailored to risks and evidence standards specific to the deployment domain. &
\cite{ouyang2021ai, mohseni2022taxonomy,  tocchetti2025ai, lefcourt2022robustness, gupta2025monitoring, roy2024enhancing, oala2020ml4h, muller2025practical, javed2024robustness, mokander2024auditing, freiesleben2023beyond, khinvasara2024robustness, balendran2025scoping, hardy2025more, casper2024black, wang2026survey, chen2025ai, petersen2022responsible,kaur2022trustworthy} \\
\midrule

MR-14.2 & Task and Model Versatility &
Robustness assessment should support the task types, model classes, modalities, datasets, and deployment settings covered by the audit scope. &
\cite{balendran2025scoping, kemmerzell2025towards, agarwal2025advancing, thelisson2024conformity, javed2024robustness, li2023trustworthy} \\
\midrule

\multicolumn{4}{l}{\textbf{Category 15: Accessibility}} \\
\midrule

MR-15.1 & Accessible and User-Friendly Robustness Tools, Visualisations, and Documents &
Robustness tools, visualisations, documents, and outputs should be understandable, usable, and available at an appropriate level of technical detail for their intended technical and non-technical audiences. &
\cite{hancox2020robustness, methnani2024s, kemmerzell2025towards, chander2025toward, li2025artificial, ojewale2025towards, agarwal2025advancing, mccormack2025trust, chen2023ai, madaio2022assessing, yu2026survey, freiesleben2023beyond, kokina2025challenges, nolte2025, javed2024robustness, li2023trustworthy, costanza2022audits} \\

\midrule

\multicolumn{4}{l}{\textbf{Category 16: Configurability}} \\
\midrule

MR-16.1 & Configurable Robustness Assessment Tools &
Robustness assessment tools should allow configuration of perturbations, thresholds, metrics, dashboards, reports, tools, and deployment-specific scenarios. &
\cite{agarwal2025advancing,  vadlapati2026ai, palazzolo2024measurable,  yang2020closer, freiesleben2023beyond, kokina2025challenges, javed2024robustness, liu2022trustworthy} \\
\midrule

\end{longtable}

\noindent\footnotesize\textit{Note.}
$^{*}$MR-8.4 was not identified during the initial synthesis of the 74 retained studies; it emerged from practitioner feedback and was subsequently supported by evidence found on re-examining those studies.
}

\subsubsection{Qualitative Coding of Open-Text Comments}
\label{sec:qualitative_analysis}

The 586 substantive open-text comments were organised as separate records and linked to an anonymous respondent identifier (P001-P128), source category, and relevant meta-requirement. Each comment was manually
coded for its theme, function, and related meta-requirement, with supporting quotations and coder notes retained for traceability. Comment functions captured confirmations, clarifications, implementation concerns, challenges, and proposed additions, while coding decisions distinguished retention, refinement, addition, merging or splitting, and no change.  Examples of anonymised comments and their Retain, Refine, Add, and Not usable coding decisions are provided in the replication package (see the Data Availability statement).
To reduce dependence on a single interpretation, one researcher coded the comments, while a second researcher independently reviewed the coding and proposed synthesis decisions, following multi-researcher coding practices used in related practitioner studies \cite{van2019practitioners,von2025navigating}. We then reconciled the qualitative feedback with the quantitative results in Phase~4. For each meta-requirement, the quantitative support status was considered together with the related practitioner comments. Proposed changes were compared against the complete set of meta-requirements to determine whether they supported retention, required refinement, or identified a concern not already represented in the existing set. This integration informed the final practitioner-informed set of meta-requirements.



\subsection{Phase 4: Meta-Requirement Synthesis}
\label{sec:framework_synthesis}

We integrated the quantitative ratings and qualitative comments to produce the final practitioner-informed set of meta-requirements. For each literature-derived meta-requirement, we considered the quantitative support status alongside the associated qualitative feedback. Practitioner suggestions were then compared with the existing meta-requirements and relevant literature to determine whether they reinforced, refined, challenged, or extended the literature-derived set of meta-requirements. Qualitative synthesis decisions were reconciled as \emph{Retain}, \emph{Refine}, \emph{Add}, \emph{No change}, or \emph{Not usable}. We retained suggested additions only when the underlying concern was not sufficiently represented by an existing meta-requirement and was supported by relevant evidence; otherwise, we incorporated them as refinements or did not adopt them. Merge and split suggestions were similarly reviewed against neighbouring meta-requirements and supporting evidence to clarify requirement boundaries where appropriate. The synthesis retained the 43 literature-derived meta-requirements, refined their wording and scope where supported by practitioner feedback and the evidence base, and added one practitioner-informed requirement. The final set therefore comprises 44 meta-requirements across the original 16 categories. 

MR-8.4, \emph{External Component and Supply Chain Assurance}, was added to Category 8, Workflow Integration, because two overall survey comments (P079, P102) identified a concern not covered by the existing meta-requirements. Comparison against the complete set showed that MR-8.1-MR-8.3 were the closest matches, but they did not address limited auditor access to external dependencies, the use of third-party validation evidence, or how these limitations affect the conclusions that can be drawn. MR-8.4 was therefore retained as a separate meta-requirement and placed in Workflow Integration because these external dependencies form part of the AI-enabled system's operational workflow. As MR-8.4 was introduced after the survey, it was not rated in the original Likert-scale validation and therefore does not share the quantitative validation status of the other 43 meta-requirements.

\section{Meta-Requirements for AI Robustness Auditing}
\label{sec:rq1_results}

This section presents the final practitioner-informed meta-requirements for AI robustness auditing (RQ1). As shown in Figure~\ref{fig:meta_requirements_framework}, the final set consists of 44 meta-requirements across 16 categories and three higher-level dimensions: Human (16 meta-requirements), Functional (23), and Quality (5). We derived the initial set of 43 meta-requirements from the 74 retained studies and refined it with practitioner feedback, including the addition of MR-8.4, \emph{External Component and Supply-Chain Assurance}. The replication package (see the Data Availability statement) reports the original 43 literature-derived meta-requirements used in the survey. The human dimension covers the organisational, stakeholder, conceptual, and competence conditions needed to make robustness auditing governable and interpretable.  

The quality dimension covers the qualities that make robustness auditing feasible, context-sensitive, understandable, usable, and configurable across different systems and deployment settings. Finally, the functional dimension covers the audit activities and evidence needed across the AI-enabled system lifecycle, including data, model, output, workflow, human review, evaluation, monitoring, and communication.  The rest of this section presents each meta-requirement and category. 


\begin{figure*}[!htbp]
\centering
\includegraphics[width=\textwidth]{ai_robustness_auditing_meta_requirements_framework.pdf}
\caption{Final practitioner-informed meta-requirements for AI robustness auditing, comprising 44 meta-requirements across 16 categories and three dimensions.}
\label{fig:meta_requirements_framework}
\end{figure*}

%



\subsection{Category 1: Standards and Governance}
\label{sec:cat1}

This category defines the measurement conventions, terminology, tools, guidance, governance responsibilities, access conditions, and audit procedures needed to make AI robustness audit evidence comparable, interpretable, and governable.

\subsubsection{MR-1.1: Standardised Metrics}
\label{mr:1.1}

Standardisation means using the same measures and the same rules for deciding whether a result is acceptable when similar robustness claims are assessed. Audit results are difficult to compare when similar claims are tested using different measures, test conditions, or acceptance rules \cite{nolte2025,blood2023reliability,tocchetti2025ai}. Prior studies measure robustness through performance under corrupted data, adversarial attacks, and changes in the data distribution \cite{oala2020ml4h,yang2020closer,nanda2021fairness,chen2023ai,khinvasara2024robustness,casper2024black,thelisson2024conformity,raji2020closing,costanza2022audits}. Each audit should therefore state what is measured, how it is measured, the conditions under which it is tested, and what result is considered acceptable \cite{madaio2022assessing,yu2026survey,nolte2025,blood2023reliability,agarwal2025advancing}.

\subsubsection{MR-1.2: Standardised Lexicon and Terminology}
\label{mr:1.2}

A common set of definitions helps everyone involved in the audit use key robustness terms consistently. This is necessary because terms such as robustness, reliability, failure, and auditability can have different meanings in engineering, legal, and organisational work \cite{tocchetti2025ai,mohseni2022taxonomy,vadlapati2026ai,freiesleben2023beyond,hornischer2026robustness,yu2026survey,wei2025trustworthy}. If these terms are not defined clearly, auditors, developers, regulators, and system users may interpret the same robustness claim differently \cite{raji2020closing,thelisson2024conformity,costanza2022audits}. The audit should therefore define the terms needed to understand the claim, the assessment, and the conclusion, and should distinguish robustness from related properties such as safety, fairness, privacy, explainability, and accountability \cite{chen2023ai,yang2020closer,nolte2025,mohseni2022taxonomy,vadlapati2026ai,ge2024survey,liu2022trustworthy,li2023trustworthy,khinvasara2024robustness,nanda2021fairness}.

\subsubsection{MR-1.3: Standardised Tools}
\label{mr:1.3}

This MR emphasizes using software tools or test instruments in a consistent way to assess the same type of robustness problem. Different tools may be used for the same type of assessment, provided that they conform to a common standard that supports consistent and comparable results  \cite{raji2020closing,ojewale2025towards,momcilovic2026assuring,oala2020ml4h,dev2025building,yu2026survey,agarwal2025advancing}. Tools differ in what they test, the conditions they can reproduce, and how they are configured, so a tool should be selected to match the robustness claim, the system or component being assessed, and the type of input data \cite{casper2024black,ge2024survey,yang2020closer,tocchetti2025ai,javed2024robustness,khinvasara2024robustness}. The audit should state which tool and version were used, the main settings applied, and any known limitations that affect the result \cite{casper2024black,kokina2025challenges,thelisson2024conformity}.

\subsubsection{MR-1.4: Actionable Guidance}
\label{mr:1.4}

This MR explains how a robustness requirement is turned into specific tasks that an auditor can carry out for the system being assessed \cite{kokina2025challenges,khinvasara2024robustness}. The guidance should translate applicable technical, legal, and organisational requirements, including requirements related to properties such as safety, security, or reliability where they are relevant, together with conditions that could cause the system to fail or be deliberately manipulated, into concrete audit actions \cite{cooper2022accountability,habbal2024artificial,thelisson2024conformity,xiong2022towards,narula2025exploring}. It should state which checks or tests to perform, what evidence to examine, and when further review is required \cite{blood2023reliability,mccormack2025trust}.

\subsubsection{MR-1.5: Robustness-aware Integration Guidance}
\label{mr:1.5}

The guidance on when and how robustness checks should be included as an AI-enabled system is developed, deployed, used, changed, and reviewed, rather than being performed only as a separate test \cite{chen2023ai,raji2020closing,thelisson2024conformity,li2023trustworthy,yu2026survey,tocchetti2025ai,madaio2022assessing}. The guidance should state at which stages the checks are required and when they must be repeated after changes to the model, data, operating conditions, or intended use of the AI-enabled system \cite{oala2020ml4h,nolte2025,khinvasara2024robustness,kokina2025challenges,wei2025trustworthy,hornischer2026robustness,javed2024robustness,liu2022trustworthy}. MR-1.5 concerns when and where robustness checks should be required across the AI lifecycle, whereas MR-8.1 concerns whether those checks are integrated into development workflows.

\subsubsection{MR-1.6: Governance Protocols}
\label{mr:1.6}

These protocols define who is responsible for investigating robustness problems, deciding what action is needed, and deciding whether the AI-enabled system may continue to be used when a problem cannot be fully resolved \cite{li2025artificial,cooper2022accountability,mccormack2025trust}. Prior work assigns these responsibilities to organisations that develop, operate, or oversee AI-enabled systems \cite{habbal2024artificial,liu2022trustworthy,raji2020closing,mokander2024auditing}. The protocols should state who investigates the problem, who decides on corrective action, who approves continued use of the system, and who must be informed when the problem could affect important properties of the AI-enabled system or its outcomes, such as safety, security, fairness, reliability, or privacy, or could result in non-compliance with applicable law \cite{laux2024trustworthy,nolte2025,thelisson2024conformity,costanza2022audits,mokander2024auditing,chamola2023review,liu2022trustworthy}.

\subsubsection{MR-1.7: Audit Access Policy}
\label{mr:1.7}

This policy defines what information and system components the auditor is allowed to inspect and what remains unavailable. Audit conclusions may depend on access to data, source code, model information, system documentation, logs, test results, and deployment information \cite{casper2024black,ojewale2025towards,koshiyama2024towards,raji2020closing,li2025artificial,chander2025toward,muller2025practical}. The policy should state what can and cannot be inspected, how long that access is available, and how any restrictions limit the conclusions that can be drawn \cite{koshiyama2024towards,simbeck2024they,thelisson2024conformity,costanza2022audits,muller2025practical,ojewale2025towards}. When privacy, security, legal, or confidentiality requirements prevent access to some information, the policy should identify the restriction and any permitted alternative source of information \cite{simbeck2024they,thelisson2024conformity,chander2025toward}.

\subsubsection{MR-1.8: Auditor Protocols}
\label{mr:1.8}

These protocols define how a robustness audit is carried out, including who performs it, the steps followed, the methods used, and how the results from those methods are used to reach the final conclusion. Audits may be conducted by internal teams or independent external assessors and may include technical testing, document review, or examination of how the organisation develops and operates the system \cite{casper2024black,thelisson2024conformity,ge2024survey,kokina2025challenges,chen2023ai}. The protocols should state the audit objective, the order of the audit activities, the methods and criteria used, and what each method contributes to the final conclusion \cite{ojewale2025towards,oala2020ml4h,costanza2022audits,casper2024black}. When more than one method is used, the protocol should explain how their results are combined to support that conclusion \cite{madaio2022assessing,nanda2021fairness,liu2022trustworthy}. MR-1.8 concerns how the audit is conducted, whereas MR-2.3 concerns the role, independence, and access of an independent auditor.

\subsection{Category 2: Engagement}
\label{sec:cat2}

This category establishes which stakeholder, user, expert, and independent-auditor perspectives must inform the framing, assessment, and use of AI robustness audit evidence.

\subsubsection{MR-2.1: Diverse Stakeholders}
\label{mr:2.1}

The stakeholders include individuals and organisations that develop, operate, use, oversee, or experience consequences from decisions made with an AI-enabled system or from its failures. An audit that relies only on technical developers may miss robustness problems, harms, or practical constraints recognised by other stakeholders \cite{chander2025toward,balendran2025scoping,costanza2022audits,robertson2023angler,li2025artificial,raji2020closing,mccormack2025trust}. The audit should therefore include perspectives from the stakeholder types connected to the robustness claim, especially where the same failure may have different consequences for different people \cite{nanda2021fairness,nolte2025,liu2022trustworthy,hancox2020robustness,tocchetti2025ai,mokander2024auditing,zahid2025explainability}.

\subsubsection{MR-2.2: Stakeholder Involvement Prioritisation}
\label{mr:2.2}

This MR determines which stakeholders need more involvement in the audit, when their input is needed, and how much participation is required. Priority should depend on how directly they may experience the consequences of a system failure, the knowledge they can contribute, and the likelihood and seriousness of the failure being assessed \cite{costanza2022audits,nolte2025,madaio2022assessing,raji2020closing,thelisson2024conformity,oala2020ml4h,li2023trustworthy}. The audit should therefore state which stakeholders need one-time, repeated, or focused involvement and why, while considering available resources and any restrictions on access to the system or information \cite{raji2020closing,thelisson2024conformity,oala2020ml4h,li2023trustworthy}.

\subsubsection{MR-2.3: Independent Auditor}
\label{mr:2.3}

Auditor independence means that the person or organisation conducting the audit can reach a conclusion without pressure from parties that develop, operate, fund, or manage the system being assessed \cite{ojewale2025towards,mccormack2025trust}. External auditors may be more separate from the organisation, while internal auditors may have more detailed knowledge of the system and its development \cite{casper2024black,liu2022trustworthy,raji2020closing}. The audit should therefore state who selected and funds the auditor, whether the auditor previously helped develop or operate the system, and whether any organisational relationship could influence the audit conclusion \cite{costanza2022audits,laux2024trustworthy,thelisson2024conformity,mokander2024auditing,mccormack2025trust,li2025artificial}. Any relationship that could influence the auditor's judgement should be disclosed and managed \cite{mccormack2025trust,li2025artificial}.

\subsubsection{MR-2.4: Early-stage Involvement}
\label{mr:2.4}

This MR implies bringing the needed expertise early in planning the robustness audit, such as technical, domain, legal or compliance, and organisational expertise, before key audit decisions are made. These decisions may include which robustness problems will be examined, which parts of the AI-enabled system will be included, which tests will be performed, what evidence will be collected, and what results will be considered acceptable. This is important because robustness problems can be harder to address when the needed expertise is introduced only after the audit has already been planned or testing has begun \cite{raji2020closing}. Experts should therefore contribute while the audit scope, test plan, evidence needs, and acceptance criteria are being decided \cite{li2025artificial,mccormack2025trust,madaio2022assessing,hornischer2026robustness,cooper2022accountability}. Their early input should help ensure that the audit examines the robustness problems and system conditions that matter for the case being audited \cite{mccormack2025trust,cooper2022accountability,madaio2022assessing,li2025artificial}.

\subsection{Category 3: Robustness Definition}
\label{sec:cat3}

This category defines the robustness claim an audit will assess, the assumptions that bound that claim, and the trade-offs that affect how robustness evidence is interpreted.

\subsubsection{MR-3.1: Scoping Robustness Concerns}
\label{mr:3.1}

This MR considers which robustness problems the audit will assess, the conditions under which they will be tested, and which conditions are excluded. The audit should state whether it covers adversarial robustness, non-adversarial robustness, or both (see Table~\ref{tab:bg_concepts} for definitions) \cite{freiesleben2023beyond,tocchetti2025ai,bayram2025towards,chen2023ai,hornischer2026robustness,li2023trustworthy,liu2022trustworthy,wang2026survey,zahid2025explainability}. It should then identify the specific conditions to be tested, such as corrupted or missing data, distribution changes, poisoning, or evasion attacks \cite{blood2023reliability,narula2025exploring,wei2025trustworthy,balendran2025scoping,vadlapati2026ai,saqib2024comprehensive,javed2024robustness,koshiyama2024towards,saeed2024security,costanza2022audits,chander2025toward,mokander2024auditing,pan2023ai,tang2026advancing}.

\subsubsection{MR-3.2: Robustness Trade-off Assumptions}
\label{mr:3.2}

A robustness trade-off occurs when improving robustness changes another important property of the system, such as accuracy, computational cost, explainability, fairness, safety, and privacy. Ignoring such changes can make a robustness improvement appear beneficial even when another property becomes worse \cite{koshiyama2024towards,liu2022trustworthy,ge2024survey,nanda2021fairness}. The audit should therefore identify which properties may change, compare them before and after the method used to improve robustness, and state how any observed gain or loss affects the robustness conclusion \cite{yang2020closer,simbeck2024they,javed2024robustness,chen2025ai,bayram2025towards,tocchetti2025ai,li2023trustworthy,vadlapati2026ai,palazzolo2024measurable}.

\subsection{Category 4: Knowledge and Skills}
\label{sec:cat4}

This category defines the competence and role-specific training needed to conduct, interpret, and maintain AI robustness audits.

\subsubsection{MR-4.1: Multidisciplinary Skills}
\label{mr:4.1}

These mean having expertise from more than one field when a robustness audit requires different kinds of knowledge. This is important because assessing robustness may require knowledge of AI models, the software system in which they operate, the domain in which the system is used, and relevant security, legal, or organisational requirements \cite{li2025artificial,mokander2024auditing,xiong2022towards}. The audit should therefore involve people with the skills needed for the system and robustness problems being assessed, which may include machine learning, software engineering, cybersecurity, statistics, law or policy, auditing, and domain expertise \cite{khinvasara2024robustness,nolte2025,laux2024trustworthy,costanza2022audits}. If required expertise is missing, the audit should state how that limitation affects the conclusions that can be drawn \cite{ala2024trusting,muller2025practical,vadlapati2026ai,kokina2025challenges,mccormack2025trust,silva2023case}.

\subsubsection{MR-4.2: Robustness-awareness Educational Programmes and Training}
\label{mr:4.2}

Training should give each person the knowledge needed for their role in robustness auditing. Auditors, developers, system administrators, users, and domain experts may need different training because they perform different tasks and make different decisions \cite{zahid2025explainability,chander2025toward}. The training should cover the robustness problems, methods, tools, and responsibilities relevant to each role, including adversarial testing or secure system operation where needed \cite{muller2025practical,kokina2025challenges,geleta2023maestro,ala2024trusting,koshiyama2024towards,habbal2024artificial}. Because methods, tools, attacks, and legal requirements change over time, the training should be updated periodically and may include practical exercises, simulated failures, or audit examples \cite{muller2025practical,zahid2025explainability}.

\subsection{Category 5: Data Auditability}
\label{sec:cat5}

This category establishes whether data used in robustness audits are sufficiently covered, documented, reliable, bias-aware, and prepared for data-level mitigation before interpreting robustness claims.

\subsubsection{MR-5.1: Data Coverage and Sufficiency}
\label{mr:5.1}

Data coverage means that the data used in the audit include the people, cases, and operating conditions covered by the robustness claim. Beyond including these cases and conditions, the audit also needs enough data to assess them reliably; this is what we mean by data sufficiency. Robustness evidence can be misleading when important groups, rare cases, distribution changes, or failure conditions are missing, or when too little data are available to assess them reliably \cite{javed2024robustness,madaio2022assessing,nanda2021fairness}. The audit should therefore check whether the available data represent the intended users, important variations in the data, rare but important cases, and the conditions under which the AI-enabled system is expected to operate, and whether enough data are available to evaluate these reliably \cite{yang2020closer,tocchetti2025ai,freiesleben2023beyond,chen2023ai,nolte2025,hornischer2026robustness,yu2026survey,oala2020ml4h,costanza2022audits,agarwal2025advancing,li2023trustworthy,wei2025trustworthy}. If important cases or conditions are missing, or are represented by too little data, the audit should state how these limitations affect the robustness conclusion \cite{raji2020closing,kokina2025challenges}.

\subsubsection{MR-5.2: Data Description, Provenance, and Lineage}
\label{mr:5.2}

This MR describes what the data contain, where they came from, and how they were collected or changed before being used in the audit. Without this information, auditors may not know whether the history of the data influenced the robustness result \cite{raji2020closing,thelisson2024conformity}. The audit should identify the data source, collection conditions, labelling process, preprocessing steps, dataset versions, and any synthetic data used \cite{oala2020ml4h,bayram2025towards,lefcourt2022robustness}. It should also trace major changes made to the data used in the audit, such as filtering, merging, relabelling, or transformation, so that their origin and processing history can be checked \cite{oala2020ml4h,costanza2022audits,ge2024survey,kokina2025challenges,raji2020closing,thelisson2024conformity}.

\subsubsection{MR-5.3: Data Quality}
\label{mr:5.3}

This concerns whether the data used in the audit are accurate, consistent, and suitable for producing reliable robustness results. Errors such as incorrect labels, duplicated records, corrupted inputs, missing values, data leakage, or contamination can distort the assessment \cite{bayram2025towards,hardy2025more,silva2023case}. The audit should therefore check for these problems and explain how any detected defects affect the robustness conclusion \cite{martinez2022software,blood2023reliability,mohseni2022taxonomy,mccormack2025trust,yang2020closer,hardy2025more,narula2025exploring,ouyang2021ai}. Where reference labels are used, their accuracy and completeness should also be checked.

\subsubsection{MR-5.4: Data Subgroup and Bias Source Identification}
\label{mr:5.4}

This MR examines whether robustness differs across defined groups in the data and whether those differences can be linked to how the data were collected, labelled, or processed. Overall results can hide weaker performance for particular demographic, geographic, device, usage, or domain groups \cite{roy2024enhancing,oala2020ml4h,li2023trustworthy,mccormack2025trust,mokander2024auditing,kemmerzell2025towards,alzubaidi2023towards}. The audit should therefore compare robustness across the groups relevant to the claim and investigate whether observed differences are associated with sampling imbalance, annotation errors, preprocessing choices, missing group information, or changes in the data distribution \cite{kemmerzell2025towards,kokina2025challenges,khinvasara2024robustness}. These data-related causes should be considered when interpreting differences in robustness across groups \cite{roy2024enhancing,alzubaidi2023towards}.

\subsubsection{MR-5.5: Robustness Pre-processing Mitigation}
\label{mr:5.5}

Pre-processing robustness mitigation refers to data-level changes made before model training or evaluation to address an identified robustness problem. Examples include cleaning corrupted data, filtering harmful inputs, transforming features, balancing data, or generating additional test or training examples \cite{dev2025building,palazzolo2024measurable,wei2025trustworthy,chen2023ai,tocchetti2025ai,freiesleben2023beyond}. The audit should assess whether the chosen data changes resolve the identified problem and whether they actually improve robustness under the conditions being tested \cite{dev2025building,javed2024robustness,tocchetti2025ai}. It should also check whether those changes hide existing weaknesses or introduce new ones \cite{wei2025trustworthy}.

\subsection{Category 6: Model Auditability}
\label{sec:cat6}

This category establishes how model behaviour, intended use, explanations, failure sources, and training-time mitigation should be made auditable for robustness claims.

\subsubsection{MR-6.1: Model Robustness Probing}
\label{mr:6.1}

This MR means deliberately testing the model under the robustness conditions selected in MR-3.1 to see how its behaviour changes. This is needed because a model may appear to perform well overall while still failing under specific conditions \cite{vadlapati2026ai,robertson2023angler,martinez2022software,hardy2025more,kaya2026explainable,petersen2022responsible}. The audit should use tests that match the robustness problem being assessed, such as corrupted inputs, distribution changes, specially constructed test cases, or adversarial examples \cite{oala2020ml4h,yang2020closer,lee2023adversarial,nolte2025,wang2026survey,tocchetti2025ai,agarwal2025advancing,ge2024survey,dev2025building,tsingenopoulos2025adaptive}. For each test, the audit should state the test conditions, the measure used, and how the result is interpreted \cite{ouyang2021ai,mokander2024auditing}.

\subsubsection{MR-6.2: Model Intended Use}
\label{mr:6.2}

This defines what the model is designed to do, who is expected to use it, and the operating conditions under which its robustness is assessed. Robustness results can be misleading if they are applied to uses or conditions that were not examined \cite{madaio2022assessing,oala2020ml4h,raji2020closing,thelisson2024conformity}. The audit should therefore state the model's intended task, expected users, role in decision-making, operating conditions, known limitations, and uses that fall outside the assessment \cite{nolte2025,madaio2022assessing,raji2020closing}. This keeps the robustness conclusion limited to the uses and conditions that were actually evaluated \cite{oala2020ml4h,thelisson2024conformity}.

\subsubsection{MR-6.3: Model Explainability}
\label{mr:6.3}

This MR encompasses using methods that help auditors understand why a model produced a particular result or behaved differently under a robustness test. This is useful when the model's behaviour cannot be understood from performance results alone \cite{khinvasara2024robustness,mccormack2025trust,mohseni2022taxonomy}. Such methods may show which inputs or model features influenced the result and may help reveal errors, bias, or unusual behaviour \cite{javed2024robustness,methnani2024s,tang2026advancing,ouyang2021ai,tocchetti2025ai,liu2022trustworthy,chander2025toward,zahid2025explainability,habbal2024artificial}. The audit should use methods that show why the model produced a result only when they help answer the robustness question being assessed, and should state what those methods can and cannot show \cite{kemmerzell2025towards,li2025artificial,kaya2026explainable,li2023trustworthy,oala2020ml4h,pan2023ai,casper2024black,chamola2023review,cooper2022accountability,ala2024trusting,kokina2025challenges,koshiyama2024towards}. Because such explanations can be unstable or misleading, they should not be treated as proof of the true cause of a model failure \cite{hancox2020robustness,kaya2026explainable,saqib2024comprehensive,ge2024survey,taesiri2022visual,casper2024black}.

\subsubsection{MR-6.4: Model Bias and Failure Source Identification}
\label{mr:6.4}

This examines which parts of the model's learned behaviour may be responsible for an observed robustness weakness. A model may perform poorly because it relies on misleading patterns, unstable features, memorised examples, or rules that do not hold outside the data used during development \cite{freiesleben2023beyond,robertson2023angler,petersen2022responsible,casper2024black,hancox2020robustness}. The audit should therefore investigate whether the weakness can be linked to the features the model relies on, the patterns it has learned, or differences in behaviour across tested cases, rather than reporting only the performance loss \cite{lee2023adversarial,kemmerzell2025towards,robertson2023angler,hancox2020robustness}. Any claimed cause should be supported by evidence from the tests performed \cite{kemmerzell2025towards}.

\subsubsection{MR-6.5: Robustness In-processing Mitigation}
\label{mr:6.5}

This mitigation method changes how the model is trained so that it performs more reliably under the failure condition being assessed. Such changes may include training with adversarial examples, adjusting the training objective, or using methods that reduce reliance on patterns that fail when the input or operating conditions change \cite{agarwal2025advancing,javed2024robustness,wei2025trustworthy,tocchetti2025ai}. The audit should assess whether the chosen training change addresses the robustness problem being tested and whether the model performs more reliably after training \cite{agarwal2025advancing,javed2024robustness}. Any improvement should be verified through testing rather than assumed from the use of the training method alone \cite{yang2020closer}.

\subsection{Category 7: Output Auditability}
\label{sec:cat7}

This category examines how robustness audits connect observed outputs and inference behaviour to the failure sources that produced them.

\subsubsection{MR-7.1: Robustness Post-processing Mitigation}
\label{mr:7.1}

Post-processing robustness mitigation refers to measures applied after model inference to reduce the effect of an identified robustness problem. Examples include rejecting uncertain outputs, checking outputs against additional rules, routing selected cases for human review, or using a safer alternative when an output cannot be trusted \cite{yu2026survey,chen2023ai}. The audit should assess whether the selected mitigation reduces the identified robustness problem under the conditions being tested and should compare system behaviour before and after the mitigation is applied \cite{yu2026survey,chen2023ai,liu2022trustworthy,tocchetti2025ai}. If the same failures remain, the post-processing mitigation should not be treated as sufficient evidence that the robustness problem has been resolved \cite{javed2024robustness,liu2022trustworthy,tocchetti2025ai}.

\subsection{Category 8: Workflow Integration}
\label{sec:cat8}

This category establishes how robustness auditing requirements are embedded into development workflows, machine-learning operations, and existing organisational systems rather than treated as isolated assessment tasks \cite{silva2023case, bayram2025towards, tocchetti2025ai}.

\subsubsection{MR-8.1: Software Development Process Integration}
\label{mr:8.1}

This MR means building robustness checks into the activities used to develop and change the AI-enabled system, rather than performing them only at the end of development. Robustness problems can be introduced when requirements, data, models, code, or system components are changed \cite{martinez2022software,silva2023case,mohseni2022taxonomy,alzubaidi2023towards,laux2024trustworthy}. The audit should therefore check whether robustness testing is included at the development stages where such changes occur and whether testing is repeated after important changes to the model, data, code, or supporting infrastructure \cite{chen2023ai,tang2026advancing,martinez2022software,silva2023case,li2023trustworthy,blood2023reliability}.

\subsubsection{MR-8.2: MLOps Frameworks Integration}
\label{mr:8.2}

This MR empathises including robustness checks in the processes used to release, deploy, update, and operate ML models. Robustness problems can appear when models or data change after development, so these processes should not rely only on tests performed before release \cite{bayram2025towards}. The audit should check whether the operational workflow includes robustness testing before release, checks on production data, verification that protective measures still work, and clear actions when required conditions are not met \cite{dev2025building,wei2025trustworthy,narula2025exploring}. Those actions may include stopping a release, pausing operation, or returning to an earlier model version \cite{ge2024survey}.

\subsubsection{MR-8.3: Robustness-aware Integration Compatibility}
\label{mr:8.3}

Integration compatibility means that the tools used for robustness auditing can work with the system being assessed and its existing technical environment. Audit tools are less useful when they cannot connect to the system, read the required data, or run under its operating constraints \cite{thelisson2024conformity,tocchetti2025ai,martinez2022software,kokina2025challenges}. The audit should therefore check whether the selected tools work with the system's interfaces, data formats, software environment, and available computing resources \cite{martinez2022software,dev2025building,koshiyama2024towards,thelisson2024conformity}. Using audit tools should not require converting data into unsupported formats, disrupt communication with the system, cause unacceptable delays, or prevent required robustness tests from being completed \cite{khinvasara2024robustness}.

\subsubsection{MR-8.4: External Component and Supply Chain Assurance}
\label{mr:8.4}

This MR concerns models, data, software, services, or infrastructure supplied by another organisation on which the audited AI-enabled system depends. Practitioner feedback identified this as a missing MR in the initial literature-based MRs: P079 highlighted "third-party and supply chain risks" and restricted auditor access to external components, while P102 called for third-party validation and guidance for auditing foundation models and third-party APIs. Prior studies likewise show that third-party components, APIs, software dependencies, data pipelines, and infrastructure can introduce failures or changes that affect the behaviour of the wider AI-enabled system \cite{dev2025building,yu2026survey,martinez2022software,blood2023reliability}. They also show that the strength of an audit conclusion depends on the information and system access available to the auditor \cite{casper2024black,mokander2024auditing}. The audit should therefore identify external dependencies that can affect the robustness claim, assess how their failure or change could alter system behaviour, and state how any inability to examine them limits the conclusion that can be drawn. To support this assessment, the audit may use Software Bills of Materials (SBOMs), which document software components and dependencies, and Artificial Intelligence Bills of Materials (AIBOMs), which extend this documentation to AI-specific assets such as models, datasets, software dependencies, configurations, and provenance information \cite{nocera2025we,radanliev2026operationalising}.

\subsection{Category 9: Human-in-the-Loop}
\label{sec:cat9}

This category defines how human judgement, review, and oversight are assigned when robustness auditing decisions cannot be supported by automated procedures alone.

\subsubsection{MR-9.1: Robustness-aware Human Participation Levels}
\label{mr:9.1}

Human participation defines when a person must review a robustness-related result, what action that person may take, and whether the system can continue operating without human approval. Human review may be needed when a robustness test fails, the system repeatedly produces unreliable results, or continued operation could create serious consequences \cite{tocchetti2025ai,kou2024beyond,mccormack2025trust,chamola2023review,chen2023ai,casper2024black,kokina2025challenges}. The audit should state when human review is required and whether the reviewer may correct a result, reject a system output, approve continued operation, or stop the system from acting \cite{oala2020ml4h,chamola2023review,thelisson2024conformity,chen2023ai,casper2024black}. Different forms of human involvement should be distinguished by whether the person only reviews the system, can intervene in its operation, or has final authority over its actions \cite{habbal2024artificial}.

\subsection{Category 10: Evaluation}
\label{sec:cat10}

This category establishes how robustness conclusions are reassessed over time, how unresolved robustness work is prioritised, and how failures are validated across system layers.

\subsubsection{MR-10.1: Continuous Robustness Assessment}
\label{mr:10.1}

This MR means checking again whether an earlier robustness conclusion still holds after the system or its operating conditions change. A conclusion reached before deployment may no longer be valid after changes to the model, data, deployment environment, protective measures, or attack methods considered in the earlier assessment \cite{blood2023reliability,li2025artificial,khinvasara2024robustness}. The audit should therefore require a new assessment at planned intervals and after important changes, such as model updates, retraining, deployment changes, robustness failures, attack methods discovered or reported after the earlier assessment, or changes to measures used to reduce those failures \cite{momcilovic2026assuring,laux2024trustworthy,chen2023ai,raji2020closing,thelisson2024conformity,simbeck2024they}. Each new assessment should identify which earlier robustness conclusions may have been affected and whether they are still supported by the available evidence \cite{mokander2024auditing,madaio2022assessing,wang2026survey,mccormack2025trust,costanza2022audits,koshiyama2024towards}.

\subsubsection{MR-10.2: Robustness-aware Progress Tracking and Prioritisation}
\label{mr:10.2}

Progress tracking means keeping a clear record of each robustness problem until it has been resolved or formally accepted. Without such tracking, problems may be forgotten, repeated, or left unresolved \cite{costanza2022audits,mokander2024auditing,chen2023ai,momcilovic2026assuring}. The audit should record the problem, its current status, the action required, and the evidence used to judge whether it has been resolved \cite{momcilovic2026assuring,chen2023ai,costanza2022audits}. Problems should be addressed first when they are likely to cause more serious or repeated consequences, or affect more users \cite{yu2026survey,robertson2023angler,petersen2022responsible}. As a follow-up, the audit should test the problem again after corrective action to determine whether it still remains \cite{chen2023ai,koshiyama2024towards,bayram2025towards}.

\subsubsection{MR-10.3: Cross-Layer Failure Validation}
\label{mr:10.3}

Cross-layer failure validation means checking whether a robustness problem stems from interactions between different parts of the AI-enabled system rather than the model alone. Such problems may involve the model, software framework, external service, infrastructure, runtime process, or human interaction \cite{yu2026survey}. The audit should identify the parts involved, reproduce the suspected failure, observe what happens, and determine whether the evidence supports the proposed cause \cite{raji2020closing,blood2023reliability,dev2025building,mohseni2022taxonomy,yu2026survey}. If some interactions cannot be tested, the audit should state which ones remain unchecked \cite{yu2026survey}.

\subsection{Category 11: Monitoring}
\label{sec:cat11}

This category establishes how deployed AI-enabled systems are observed for robustness-relevant change, runtime violations, and auditable operational records after the original assessment.

\subsubsection{MR-11.1: Continuous, Automated, and Real-time Robustness Monitoring}
\label{mr:11.1}

Robustness monitoring means observing a deployed AI-enabled system for changes that may reduce its ability to behave reliably. Such changes may include worsening performance, changes in incoming data, unusual outputs, signs that the system is being deliberately manipulated, or changes in operating conditions \cite{blood2023reliability,chen2023ai,li2025artificial,gupta2025monitoring,mohseni2022taxonomy,nolte2025,thelisson2024conformity,chen2025ai,vadlapati2026ai,narula2025exploring,methnani2024s,bayram2025towards,robertson2023angler,khinvasara2024robustness}. Monitoring may be continuous, meaning that the system is checked throughout operation, and automated where software can perform those checks without repeated manual inspection. Where rapid response is required, changes should be detected and reported as they occur or soon afterwards. The audit should state what is monitored, how often it is checked, and what change should trigger further investigation or reassessment (see MR-10.1 and MR-10.2) \cite{chen2023ai,bayram2025towards,koshiyama2024towards,robertson2023angler,martinez2022software,chamola2023review,kaya2026explainable,ojewale2025towards}. Automated alerts should be checked before being treated as evidence that robustness has degraded \cite{blood2023reliability,raji2020closing,narula2025exploring,schmelczer2023trustworthy,ala2024trusting,mccormack2025trust}.

\subsubsection{MR-11.2: Robustness-aware Logging}
\label{mr:11.2}

This means keeping records that allow auditors to reconstruct what happened when a robustness-related problem occurs. Without such records, it may be difficult to determine what the system received, what it produced, which model version was running, or what changes had been made \cite{methnani2024s,li2025artificial}. The audit should therefore check whether logs capture the information needed to investigate the event, including relevant inputs, outputs, model versions, system settings, alerts, and recorded failures \cite{thelisson2024conformity,schmelczer2023trustworthy,raji2020closing,li2025artificial}. The logs should be kept for a defined period, protected from unauthorised access or alteration, and available when later investigation is required \cite{thelisson2024conformity,schmelczer2023trustworthy,li2025artificial}.

\subsection{Category 12: Communication}
\label{sec:cat12}

This category establishes how evidence from robustness audits is communicated through visual interfaces, risk reports, and user-facing documentation so the intended audience can inspect, interpret, and act on audit results.

\subsubsection{MR-12.1: Robustness Dashboards and Visualisation}
\label{mr:12.1}

This MR entails presenting robustness audit results in a visual form so that users can quickly see important robustness changes, problems, and trends. This is useful when relevant information is spread across different measures, AI-enabled system components, or periods of system operation \cite{nolte2025,li2025artificial}. The audit should check whether the visualisation clearly shows the robustness measures being tracked, important changes over time, detected problems, and any uncertainty that affects interpretation \cite{xiong2022towards,robertson2023angler,tocchetti2025ai,methnani2024s,muller2025practical,raji2020closing,ala2024trusting}. It should also identify who the visualisation is intended for and present enough detail for that audience to understand the result without hiding important problems or limitations \cite{robertson2023angler,oala2020ml4h,nolte2025,agarwal2025advancing,kokina2025challenges}.

\subsubsection{MR-12.2: Robustness Documentation and Risk Report}
\label{mr:12.2}

The written audit report provides a clear record of what was assessed, how the assessment was performed, what was found, and what limits the conclusion. Without this information, readers may not know which evidence supports the robustness conclusion or where uncertainty remains \cite{mokander2024auditing,koshiyama2024towards,laux2024trustworthy,costanza2022audits,ojewale2025towards,mccormack2025trust,petersen2022responsible,kaya2026explainable,raji2020closing,momcilovic2026assuring}. The report should state the audit objective, robustness claim, methods used, evidence examined, main findings, known limitations, unresolved problems, and any required follow-up actions. It should also identify the system, model, data, and test conditions to which the conclusion applies, drawing on retained operational records where needed (see MR-11.2) \cite{li2023trustworthy,kemmerzell2025towards,oala2020ml4h,koshiyama2024towards,palazzolo2024measurable,vadlapati2026ai,bayram2025towards,cooper2022accountability,simbeck2024they,tsingenopoulos2025adaptive,ouyang2021ai,gupta2025monitoring,blood2023reliability,yu2026survey}.

\subsection{Category 13: Efficiency}
\label{sec:cat13}

This category examines whether robustness auditing activities can be executed within financial, temporal, computational, and operational constraints without overstating what constrained evidence can support.

\subsubsection{MR-13.1: Resource Efficiency and Scalability}
\label{mr:13.1}

This MR concerns whether a robustness audit can be carried out with the available time, computing resources, infrastructure, and people with the required expertise as the size or complexity of the system increases. Audits may become difficult to perform when models, datasets, or system environments require substantial computation, specialist expertise, or long execution times \cite{kokina2025challenges,tocchetti2025ai,kou2024beyond,saqib2024comprehensive,madaio2022assessing,nanda2021fairness,khinvasara2024robustness,kaya2026explainable,thelisson2024conformity,martinez2022software,chen2025ai,dev2025building,wang2026survey,gupta2025monitoring,mccormack2025trust}. The audit should therefore state the resources and time required and whether those limits reduce the amount of testing that can be performed \cite{javed2024robustness,wei2025trustworthy,tocchetti2025ai,saqib2024comprehensive,mokander2024auditing}. If resource limits require fewer tests or smaller-scale evaluation, the robustness conclusion should be limited to what was actually assessed \cite{mohseni2022taxonomy,hornischer2026robustness,kokina2025challenges,nanda2021fairness,javed2024robustness,wei2025trustworthy}.

\subsection{Category 14: Adaptability}
\label{sec:cat14}

This category establishes whether robustness audits are adapted to the domain, task, model type, data modality, and deployment setting being assessed.

\subsubsection{MR-14.1: Domain-specific Adaptability}
\label{mr:14.1}

This MR means adjusting the robustness audit to the characteristics of the application area in which the AI-enabled system is used. A test that is suitable in one domain may be misleading in another because operating conditions, acceptable performance, and the consequences of failure can differ \cite{freiesleben2023beyond,hardy2025more}. The audit should therefore adapt the test conditions, measures, and acceptance criteria to the domain, including the type of data, expected operating conditions, and robustness problems that matter for that application \cite{oala2020ml4h,mokander2024auditing,javed2024robustness,ouyang2021ai,petersen2022responsible,chen2025ai,wang2026survey,balendran2025scoping,mohseni2022taxonomy,gupta2025monitoring}. The robustness conclusion should apply only to the domain and conditions that were actually assessed \cite{roy2024enhancing,ouyang2021ai,lefcourt2022robustness,muller2025practical,freiesleben2023beyond,tocchetti2025ai,khinvasara2024robustness,javed2024robustness,casper2024black,mokander2024auditing}.

\subsubsection{MR-14.2: Task and Model Versatility}
\label{mr:14.2}

This MR means checking whether an audit method remains suitable when the AI task, model type, or input data format changes. A method designed for one setting may give misleading results when transferred to another without adjustment \cite{kemmerzell2025towards}. The audit should therefore assess whether the chosen method fits the system's task, model type, input data, and intended use, and should state any limits on using that method in other settings \cite{balendran2025scoping,kemmerzell2025towards,li2023trustworthy}. For example, a method developed for a classification task, a neural-network model, or a single input type may not work in the same way for other tasks, model types, or systems that use text, images, audio, or combinations of these inputs \cite{balendran2025scoping}. Where no well-established assessment method exists, the audit should state that limitation rather than assume that one method is suitable for every model or task \cite{thelisson2024conformity}.

\subsection{Category 15: Accessibility}
\label{sec:cat15}

This category concerns whether tools for robustness auditing, explanations, visualisations, documents, and outputs are understandable and usable by their intended audiences \cite{hancox2020robustness, ojewale2025towards, zahid2025explainability}.

\subsubsection{MR-15.1: Accessible and User-Friendly Robustness Tools, Visualisations, and Documents}
\label{mr:15.1}

Accessibility means that tools for robustness auditing can be operated, visualisations can be understood, and documents can be read and interpreted by their intended users. User-friendly means that these materials are easy to navigate and use without unnecessary effort. Different users may have different levels of technical knowledge, language needs, or accessibility requirements \cite{hancox2020robustness,kemmerzell2025towards,chander2025toward,li2025artificial,ojewale2025towards}. The audit should therefore check whether tool interfaces are easy to use, visualisations have clear labels and explanations, and documents use language and detail suited to the intended audience \cite{chander2025toward,ojewale2025towards}. Non-technical users should receive clear summaries, while qualified users should be able to access the technical detail needed to examine the robustness evidence \cite{methnani2024s,kemmerzell2025towards,mccormack2025trust,agarwal2025advancing,chen2023ai}.

\subsection{Category 16: Configurability}
\label{sec:cat16}

This category establishes whether robustness-assessment tools can be configured to match the scoped system, domain, thresholds, perturbations, attack assumptions, and deployment conditions.

\subsubsection{MR-16.1: Configurable Robustness Assessment Tools}
\label{mr:16.1}

Configurability means that the settings of a robustness-assessment tool can be changed to match the system and the robustness condition being tested. Fixed default settings may give misleading results when the audit requires different levels of input or operating-condition change, different acceptance criteria, or different assumptions about how inputs or system conditions may be changed \cite{agarwal2025advancing,kokina2025challenges,freiesleben2023beyond,yang2020closer,javed2024robustness,palazzolo2024measurable,vadlapati2026ai,liu2022trustworthy}. The audit should therefore check whether the required settings can be changed and should state which settings were used and why. Any change from the default settings should be explained, and the selected settings should still match the conditions covered by the robustness claim \cite{freiesleben2023beyond,yang2020closer,javed2024robustness,liu2022trustworthy,kokina2025challenges,palazzolo2024measurable}.

\begin{tcolorbox}[title={Key Findings of RQ1},
colback=gray!5, colframe=gray!60]

\textbf{Finding 1.} The literature review produced 43 meta-requirements organised into 16 categories and three dimensions: Human, Functional, and Quality.

\textbf{Finding 2.} Most literature-derived meta-requirements concern audit activities and evidence (22 Functional), compared with organisational and human aspects (16 Human) and audit-approach qualities (5 Quality).

\end{tcolorbox}


\section{Practitioners' Perspectives on the Meta-Requirements}
\label{sec:rq2_results}

RQ2 examines practitioners' perspectives on the 43 literature-derived meta-requirements evaluated in the survey. The analysis draws on 5{,}504 Likert ratings from 128 valid respondents and 586 substantive comments. We report respondent characteristics, quantitative support, and qualitative feedback, including proposed refinements, challenges,
and additional concerns. These findings informed the final set of meta-requirements presented in Section~\ref{sec:rq1_results}.


\subsection{Respondent Characteristics}
\label{sec:practitioner_sample}

After screening, we retained 128 valid responses. Respondent characteristics describe the validation sample's experience and demographic composition, including AI/ML roles, AI robustness assessment or AI robustness auditing experience, robustness-related activities, AI/ML experience, location, age, gender, and education.

\subsubsection*{Professional Background and Experience}
Respondents represented diverse AI/ML-related backgrounds, including machine-learning engineers, data scientists, AI/ML auditors, software engineers, data engineers, and other related roles. Figure~\ref{fig:roles_by_robustness_experience} shows these roles together with respondents' AI robustness assessment or auditing experience. The two largest experience groups had less than one year (40\%) and one to three years (41\%) of robustness assessment or auditing experience, respectively, while smaller groups reported longer experience or no prior experience. The sample therefore represents respondents with varying levels of AI/ML and robustness-assessment experience, rather than exclusively experienced AI robustness auditors. The findings should be interpreted in light of this participant composition.

\subsubsection*{Robustness-related activities}
Respondents had participated in a range of activities directly relevant to the validation task. The most frequently reported were robustness testing (73/128, 57.03\%), failure or error analysis (71/128, 55.47\%), robustness-focused model evaluation (59/128, 46.09\%), robustness metric selection or definition (54/128, 42.19\%), and adversarial or non-adversarial evaluation (54/128, 42.19\%). Other reported activities included domain-specific robustness evaluation (46/128, 35.94\%), development or use of robustness-assessment tools (44/128, 34.38\%), AI robustness auditing or assessment (42/128, 32.81\%), and post-deployment robustness monitoring (29/128, 22.66\%). The complete distribution of respondents' robustness-related activities is provided in the replication package (see the Data Availability statement).

\subsubsection*{Geographic distribution}
\begin{figure}[htbp]
    \centering

    \begin{minipage}[t]{0.48\textwidth}
        \centering
        \includegraphics[
            width=\linewidth,
            keepaspectratio
        ]{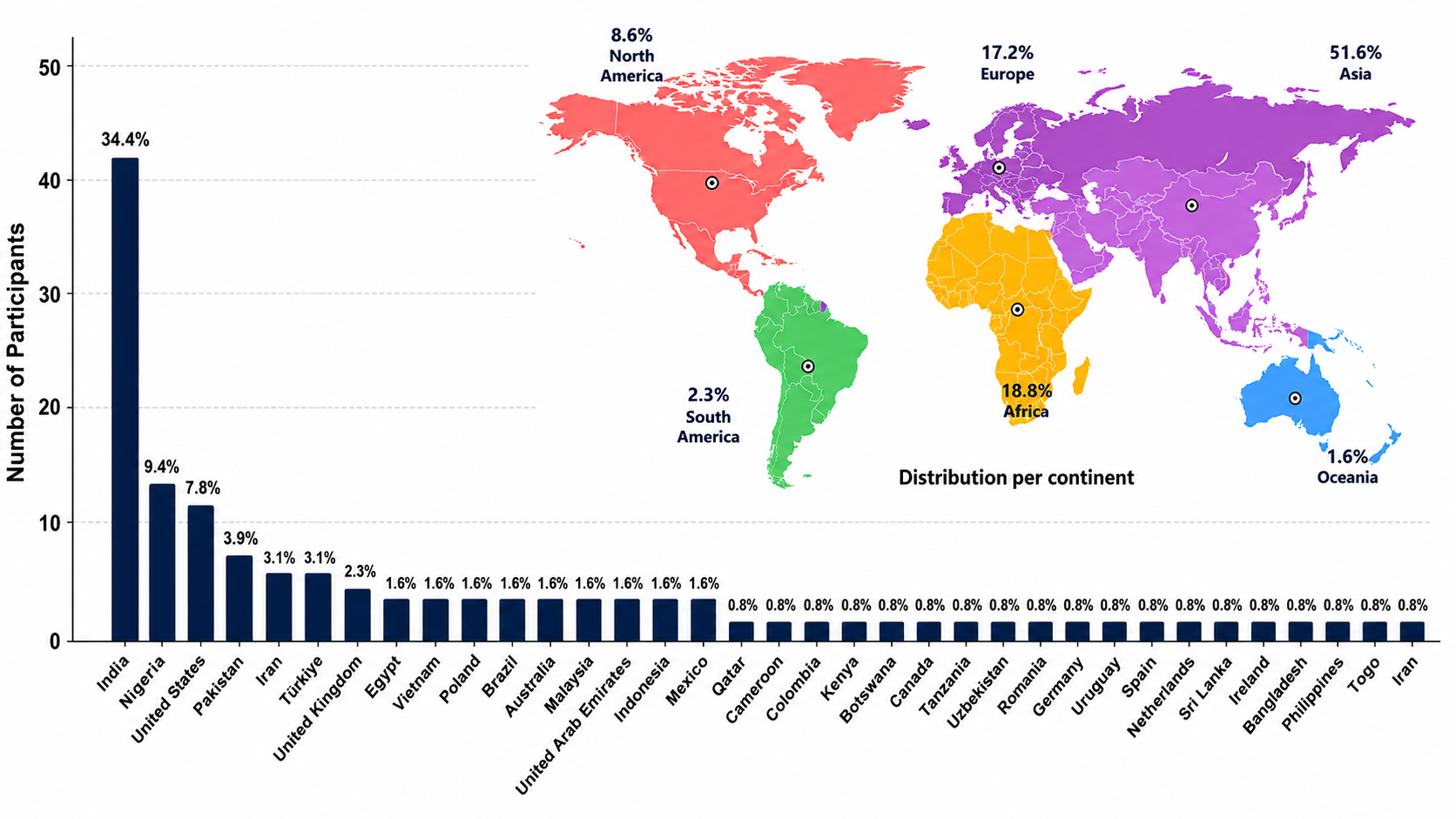}
        \caption{Geographic distribution of respondents by country and continent.}
        \label{fig:geographic_distribution}
    \end{minipage}
    \hfill
    \begin{minipage}[t]{0.48\textwidth}
        \centering
        \includegraphics[
            width=\linewidth,
            keepaspectratio
        ]{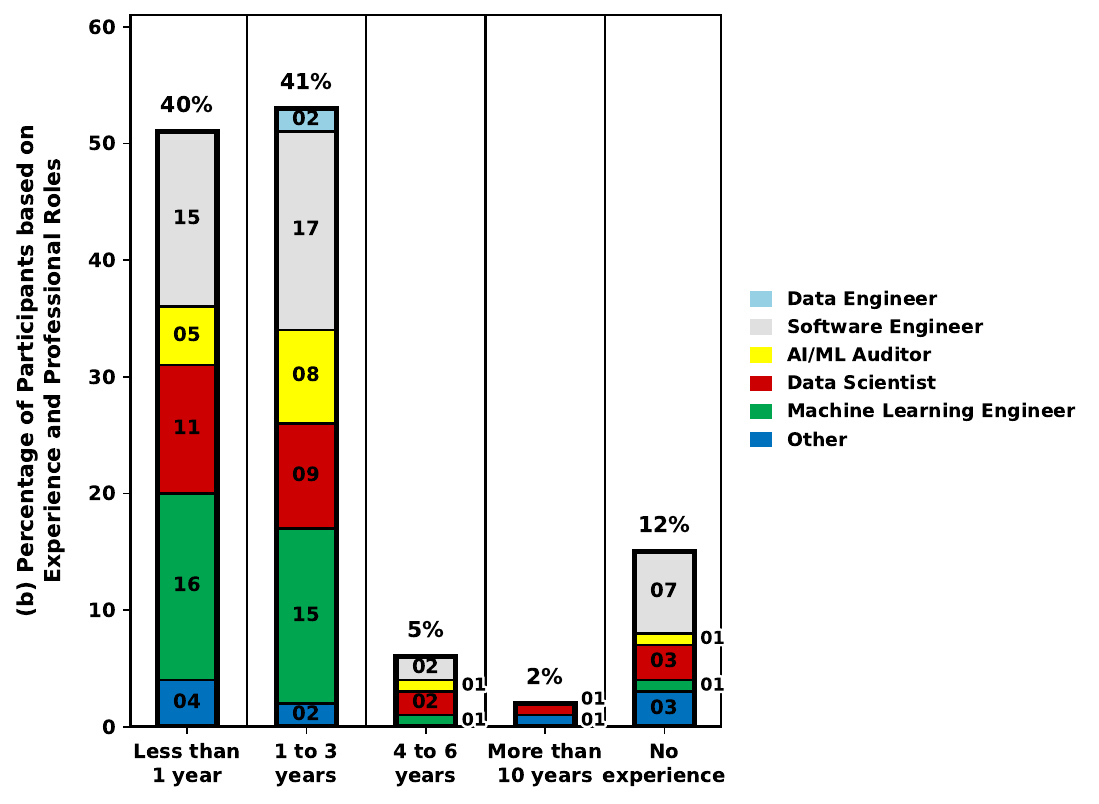}
        \caption{Distribution of participants by robustness auditing experience and primary AI/ML-related role.}
        \label{fig:roles_by_robustness_experience}
    \end{minipage}

\end{figure}

Figure~\ref{fig:geographic_distribution} shows that respondents came from multiple countries across six continents. Asia accounted for the largest share of the sample (66/128, 51.56\%), followed by Africa (24/128, 18.75\%), Europe (22/128, 17.19\%), North America (11/128, 8.59\%), South America (3/128, 2.34\%), and Oceania (2/128, 1.56\%). The largest country-level contributions came from India, Nigeria, and the United States, followed by Pakistan, Kenya, and T\"urkiye.


\subsubsection*{Demographic characteristics:}
The sample also covered different age and educational backgrounds. Most respondents were aged 18-24 years (73/128, 57.03\%), followed by 25-34 years (29/128, 22.66\%) and 35-44 years (16/128, 12.50\%). The remaining respondents were distributed across older age groups. Full distributions for age, gender, education, AI/ML experience, and robustness-related experience are provided in the replication package (see the Data Availability statement).


\subsection{Overall Quantitative Validation}
\label{sec:overall_validation_results}

The 128 valid responses yielded 5{,}504 ratings across the 43 literature-derived meta-requirements. Ratings of 4-5 were classified as agreement, 3 as neutral, and 1-2 as disagreement. Overall, 4{,}469 ratings (81.20\%) indicated agreement, 709 (12.88\%) were neutral, and 326 (5.92\%) indicated disagreement. By response option, 2{,}624 ratings (47.67\%) were \emph{strongly agree}, 1{,}845 (33.52\%) \emph{agree}, 235 (4.27\%) \emph{disagree}, and 91 (1.65\%) \emph{strongly disagree}.



\begin{table}[htbp]
\centering
\scriptsize
\setlength{\tabcolsep}{3pt}

\begin{minipage}[t]{0.32\textwidth}
\centering
\caption{Overall practitioner validation results across all MR ratings.}
\label{tab:overall_validation_summary}

\begin{tabularx}{\linewidth}{@{}Xrr@{}}
\toprule
\textbf{Response group} &
\shortstack{\textbf{No.}\\\textbf{ratings}} &
\textbf{\%} \\
\midrule
Agree / Strongly agree & 4{,}469 & 81.20 \\
Neutral & 709 & 12.88 \\
Disagree / Strongly disagree & 326 & 5.92 \\
& & \\
& & \\
\midrule
\textbf{Total} & \textbf{5{,}504} & \textbf{100.00} \\
\bottomrule
\end{tabularx}
\end{minipage}
\hfill
%
\begin{minipage}[t]{0.32\textwidth}
\centering
\caption{Distribution of individual Likert responses across all MR ratings.}
\label{tab:overall_likert_distribution}

\begin{tabularx}{\linewidth}{@{}Xrr@{}}
\toprule
\textbf{Likert response} &
\shortstack{\textbf{No.}\\\textbf{ratings}} &
\textbf{\%} \\
\midrule
Strongly agree & 2{,}624 & 47.67 \\
Agree & 1{,}845 & 33.52 \\
Neutral & 709 & 12.88 \\
Disagree & 235 & 4.27 \\
Strongly disagree & 91 & 1.65 \\
\midrule
\textbf{Total} & \textbf{5{,}504} & \textbf{100.00} \\
\bottomrule
\end{tabularx}
\end{minipage}
\hfill
%
\begin{minipage}[t]{0.32\textwidth}
\centering
\caption{Support classification of rated meta-requirements.}
\label{tab:mr_support_status}

\begin{tabularx}{\linewidth}{@{}Xr@{}}
\toprule
\textbf{Support status} &
\shortstack{\textbf{No. of}\\\textbf{MRs}} \\
\midrule
Strongly supported & 41 \\
Supported & 2 \\
Mixed or unclear & 0 \\
Challenged & 0 \\
& \\
\midrule
\textbf{Total rated MRs} & \textbf{43} \\
\bottomrule
\end{tabularx}
\end{minipage}

\end{table}

These results indicate broad practitioner support for the literature-derived meta-requirements for AI robustness auditing. However, the neutral and disagreement-level responses show that validation should not be interpreted as unqualified acceptance. Instead, the quantitative results establish the overall level of practitioner support, while the qualitative comments explain where clarification, refinement, contextualisation, or extension was needed.


The absence of a quantitatively challenged MR does not mean that practitioners raised no concerns. It means no MR reached the predefined aggregate threshold to be classified as challenged. The 326 disagreement-level ratings and the open-text comments therefore remained important for interpreting conditional acceptance, feasibility concerns, scope concerns, and overlap concerns.

\subsection{Meta-Requirement-Level Validation}
\label{sec:mr_level_validation}

Table~\ref{tab:mr_validation_full} summarises the practitioner ratings for the 43 literature-derived meta-requirements. Overall, support was consistently high: agreement or strong agreement ranged from 71.88\% to 88.28\% across the individual meta-requirements. Of the 43 meta-requirements, 41 were classified as strongly supported and two as supported; none was quantitatively classified as challenged.

MR-8.3, concerning compatibility with AI-enabled system infrastructure, received the highest agreement (88.28\%), followed by MR-5.3 on robustness-focused data quality checking (87.50\%) and MR-12.2 on documenting robustness risks, evidence, and conclusions (86.72\%). The two meta-requirements classified as supported rather than strongly supported were MR-2.2 on risk-based stakeholder involvement
(71.88\%) and MR-6.5 on robustness-focused training-time mitigation (73.44\%). Despite their comparatively lower ratings, both retained clear majority agreement.

\subsubsection{Category-Level Validation}
\label{sec:category_level_validation}

The category-level analysis assessed whether practitioner support extended across the original survey structure. All 16 categories received broad agreement-level support (Table~\ref{tab:category_agreement_ranking}).


\begin{table}[htbp]
\centering
\caption{Category-level agreement ranking across the original 16 survey categories.}
\label{tab:category_agreement_ranking}
\tiny
\setlength{\tabcolsep}{3pt}
\renewcommand{\arraystretch}{0.9}

\begin{tabular}{r l r @{\hspace{1.2em}} r l r}
\toprule
\textbf{Rank} & \textbf{Category} & \textbf{Agree (\%)} &
\textbf{Rank} & \textbf{Category} & \textbf{Agree (\%)} \\
\midrule
1  & Workflow Integration        & 86.20 & 9  & Communication            & 81.64 \\
2  & Configurability             & 83.59 & 10 & Robustness Definition     & 81.25 \\
3  & Data Auditability           & 83.12 & 11 & Human-in-the-Loop         & 81.25 \\
4  & Knowledge and Skills        & 82.81 & 12 & Accessibility             & 81.25 \\
5  & Output Auditability         & 82.81 & 13 & Efficiency                & 79.69 \\
6  & Monitoring                  & 82.81 & 14 & Standards and Governance  & 79.49 \\
7  & Adaptability                & 82.81 & 15 & Model Auditability        & 79.38 \\
8  & Evaluation                  & 82.55 & 16 & Engagement                & 76.37 \\
\bottomrule
\end{tabular}
\end{table}

Practitioner support extended across all 16 survey categories. Qualitative feedback identified needs for clearer wording and sharper boundaries between some meta-requirements. We therefore interpreted category-level ratings alongside the open-text feedback.

\subsection{Qualitative Practitioner Validation and Meta-Requirement Refinement}
\label{sec:qualitative_practitioner_validation}

Open-text responses complemented the quantitative ratings by showing how practitioners proposed retaining, refining, challenging, or extending the literature-derived meta-requirements. Comments were coded (open coding) as confirmations, refinements, scope clarifications, implementation concerns, additions, merge/split suggestions, or challenges. During reconciliation, we compared each proposed change with the existing meta-requirements and other comments; we treated proposals already covered as refinements rather than new additions, merges, or splits. The final decisions comprised 162 retentions, 394 refinements, 13 no-change cases, and 15 comments not usable for meta-requirement synthesis. Only two comments retained an \emph{Add} decision, both identifying the same previously uncovered concern.

\subsubsection*{Retention of the literature-derived meta-requirements.}
Practitioner comments reinforced the broad quantitative support for the literature-derived meta-requirements by explaining why specific requirements remained important. P003 described governance and lifecycle requirements as "essential and foundational for credible audits," while P011 and P034 emphasised continuous and proactive lifecycle integration. P117 similarly highlighted standardised metrics, terminology, governance, and audit procedures as important for consistent, reproducible, and comparable assessments. Comparable confirmations across data, model, output, monitoring, adaptability, and configurability concerns (e.g., P019, P031, P047, P072, P091, P110, P117) supported retention of the underlying requirements while allowing further refinement of their scope and implementation.

\subsubsection*{Refinement of existing meta-requirements}
Refinement was the dominant qualitative outcome. Practitioners generally accepted the literature-derived requirements but requested clearer boundaries, applicability conditions, evidence expectations, and implementation guidance. A recurring concern was balancing standardisation with contextual flexibility (P006, P010, P012, P016, P019, P040, P096, P121, P123). Respondents supported common metrics, terminology, governance, and reporting expectations, while cautioning against requiring identical tools or methods across contexts. P016 explicitly called for a "balance [between] standardization and flexibility," and P123 argued that standardisation should support comparability without constraining new evaluation methods, while also noting an "absence of actionable, tool-agnostic guidance." These comments helped distinguish common expectations for comparability from justified, context-specific choices of tools, metrics, and evaluation methods.

Practitioners also refined audit scope, access conditions, and assurance claims (P002, P006, P017, P024, P092, P112, P123). P002 called for explicit "out-of-scope" boundaries, while P017 and P112 distinguished white-, grey-, and black-box contexts and argued that audit requirements should reflect available access. P123 likewise identified "unclear access scope" as a practical barrier, and P024 and P092 requested clearer assumptions, thresholds, limitations, and evidence conditions. These comments led to making access, exclusions, assumptions, and limitations explicit and aligning assurance claims with available evidence.

Risk proportionality and lifecycle reassessment formed a second cross-cutting pattern (P005, P016, P019, P024, P047, P048, P072, P077, P081, P092, P106, P108, P121). Practitioners argued that stakeholder involvement, review, testing, monitoring, and reassessment should vary with system risk, deployment context, and material change. P121 cautioned against "over-burdening low risk systems," while P024 proposed risk-tiered requirements. P047 stressed that robustness is valid only for a specific context and time, and P058 described it as a "living process." P077 and P081 further supported change-sensitive reassessment through scheduled, event-driven, and incident-driven triggers. These comments refined continuous assessment into a risk-sensitive, change-triggered process rather than repeated testing after every change.

Stakeholder participation, expertise, independence, and human oversight were also refined (P002, P005, P006, P010, P016, P017, P022, P047, P057, P091, P117, P122, P125). P006 called for documenting stakeholder input and resolving conflicting viewpoints, while P057 distinguished stakeholder identification from risk-based prioritisation. P047 argued that domain expertise should shape the test basis, assumptions, risks, and evidence needs. P002 and P057 further stressed sufficient technical access and independence from development or commercial pressures. For human oversight, P002 warned against "rubber-stamping," while P005, P016, and P091 linked involvement to risk, reviewer expertise, defined intervention conditions, and auditable decisions. These comments clarified who should participate, when their input is required, and how to exercise independence and oversight.

A second refinement concerned the distinction between auditing and mitigation and the evidence needed to support conclusions (P034, P037, P047, P058, P072, P091, P112, P123). Practitioners stressed that the presence of a mitigation does not establish robustness; audits should evaluate its effectiveness, conditions, limitations, and trade-offs. P058 warned against "a conflation of auditing with engineering," while P112 called for assessment of the "actual impact and trade-offs" of mitigations. P034, P072, and P091 further emphasised testing mitigations across failure types, stress conditions, users, and deployment scenarios. Related comments requested clearer thresholds, versioning, reproducibility, reporting, scalable testing, and configuration records (e.g., P024, P047, P072, P091, P112, P123). We incorporated these points as refinements where the underlying requirement already existed.

\subsubsection*{Challenges and qualifications.}
No literature-derived meta-requirement was removed; however, some practitioner comments questioned whether specific implementation mechanisms should be universal or mandatory. The clearest qualifications concerned workflow integration and communication. P121 described workflow and MLOps integration as "important but not foundational," while P123 questioned treating mature MLOps integration as a prerequisite for valid auditing. P108 and P125 likewise argued that integration depth should depend on system complexity, deployment context, risk, resources, and MLOps maturity. P077 viewed dashboards as optional, whereas P123 and P113 stressed that dashboards should not replace a documented and authoritative evidence base. These comments challenged implementation-specific mechanisms rather than the underlying needs for integration, communication, monitoring, and evidence retention. Other comments qualified how evidence should be interpreted. P058 supported stakeholder involvement but argued that audit conclusions should remain evidence-driven. P077 also questioned whether accessibility should have the same mandatory status as other meta-requirements, although other respondents emphasised accessible, audience-appropriate communication (e.g., P037, P047, P054, P072, P081, P091, P097, P099). These comments were retained as qualifying evidence and used to introduce proportionality, conditionality, and clearer distinctions between required audit outcomes and optional implementation mechanisms.

\subsubsection*{Practitioner-identified extension.}
Several comments initially proposed additional requirements, including reproducibility and re-audit, auditor independence, stakeholder-conflict management, competency assessment, risk-tiered assurance, incident response, evidence sufficiency, configuration governance, and regulatory or reporting controls (e.g., P006, P022, P024, P058, P081, P091, P122, P125). Reconciliation against the literature-derived meta-requirements showed that most clarified or operationalised existing requirements and were therefore incorporated as refinements. One concern remained insufficiently represented: assurance for externally controlled components and dependencies. P079 called for addressing "third-party and supply chain risks," including external models, foundation-model APIs, pretrained components, datasets, libraries, plugins, and cloud services, while noting that restricted visibility limits defensible assurance claims. P102 independently highlighted third-party validation and auditing of foundation models and APIs, with related concerns also raised by P111 and P123 (P079, P102, P111, P123). Because P079 and P102 identified the same uncovered concern, their convergent evidence supported one additional practitioner-informed meta-requirement, MR-8.4, \emph{External Component and Supply Chain Assurance}.

\subsubsection*{Merge and split proposals.}
Practitioners identified overlap or uneven granularity among some literature-derived requirements (P034, P057, P081, P091, P092, P113). P034 noted overlap among MR-1.1, MR-1.2, and MR-1.3, while P057 distinguished stakeholder identification (MR-2.1) from risk-based prioritisation (MR-2.2). P113 similarly observed overlap between MR-6.1 and MR-6.5 but supported separating stress-test outcomes from the rationale and effectiveness of training-time mitigation. P081, supported by P037 and P091, proposed distinguishing output-failure identification from post-training mitigation assessment. Broader comments from P091 and P092 also noted overlap across evaluation, monitoring, workflow integration, documentation, accessibility, adaptability, and configurability. No MR-level merge or split was adopted; these comments were used to sharpen requirement boundaries, while the original 16-category structure was retained.

\subsubsection*{Integrated effect of practitioner validation.}
Practitioner validation was primarily refinement-oriented rather than replacement-oriented. The literature-derived concerns were broadly retained, while practitioners clarified their scope, applicability, evidence expectations, and implementation. Challenges mainly targeted universal or implementation-specific interpretations, and proposed merges and splits were addressed through boundary clarification or refinement of existing requirements. The main exception was the gap concerning external components and supply chain dependencies, which resulted in one new meta-requirement. Overall, the 43 literature-derived meta-requirements were refined and extended to a practitioner-informed set of 44 meta-requirements while retaining the 16-category structure.

\begin{tcolorbox}[title={Key Findings of RQ2},
colback=gray!5, colframe=gray!60]

\textbf{Finding 3.} Practitioner ratings showed broad support for the literature-derived meta-requirements: 41 of the 43 meta-requirements were strongly supported and two were supported, with MR-level agreement ranging from 71.88\% to 88.28\%.

\textbf{Finding 4.} Practitioner feedback primarily refined the existing meta-requirements and identified one additional standalone requirement on external-dependency robustness, resulting in a final set of 44 meta-requirements across 16 categories.

\end{tcolorbox}

\begin{center}
\tiny
\setlength{\tabcolsep}{2.6pt}
\renewcommand{\arraystretch}{1.0}

\begin{longtable}{
c
c
p{6.2cm}
c
c
c
c
c
c
c
c
c
p{2.2cm}
}
\caption{Practitioner validation results for the 43 literature-derived meta-requirements ($n=128$ respondents per meta-requirement).}
\label{tab:mr_validation_full}\\

\toprule
\textbf{Cat.} &
\textbf{MR} &
\textbf{Meta-requirement} &
\textbf{$n$} &
\textbf{Mean} &
\textbf{SD} &
\textbf{Median} &
\textbf{Mode} &
\textbf{IQR} &
\textbf{Agree (\%)} &
\textbf{Neutral (\%)} &
\textbf{Disagree (\%)} &
\textbf{Quantitative status} \\
\midrule
\endfirsthead

\multicolumn{13}{c}%
{{\tablename\ \thetable{} - continued from previous page}}\\
\toprule
\textbf{Cat.} &
\textbf{MR} &
\textbf{Meta-requirement} &
\textbf{$n$} &
\textbf{Mean} &
\textbf{SD} &
\textbf{Median} &
\textbf{Mode} &
\textbf{IQR} &
\textbf{Agree (\%)} &
\textbf{Neutral (\%)} &
\textbf{Disagree (\%)} &
\textbf{Quantitative status} \\
\midrule
\endhead

\midrule
\multicolumn{13}{r}{Continued on next page}\\
\endfoot

\bottomrule
\endlastfoot

1 & MR-1.1 &
Should use standardized robustness metrics &
128 & 4.23 & 1.12 & 5.0 & 5 & 1.0 &
82.81 & 6.25 & 10.94 &
Strongly supported \\

1 & MR-1.2 &
Should use shared robustness terminology &
128 & 4.15 & 1.04 & 4.0 & 5 & 1.0 &
76.56 & 16.41 & 7.03 &
Strongly supported \\

1 & MR-1.3 &
Should use standardized robustness tools &
128 & 3.90 & 0.92 & 4.0 & 4 & 0.0 &
78.12 & 12.50 & 9.38 &
Strongly supported \\

1 & MR-1.4 &
Should provide actionable audit guidance &
128 & 4.16 & 1.00 & 4.0 & 5 & 1.0 &
75.78 & 17.19 & 7.03 &
Strongly supported \\

1 & MR-1.5 &
Should integrate robustness into AI lifecycle workflows &
128 & 4.30 & 0.93 & 5.0 & 5 & 1.0 &
81.25 & 14.06 & 4.69 &
Strongly supported \\

1 & MR-1.6 &
Should establish robustness governance protocols &
128 & 4.31 & 0.97 & 5.0 & 5 & 1.0 &
83.59 & 10.16 & 6.25 &
Strongly supported \\

1 & MR-1.7 &
Should define audit access requirements &
128 & 4.22 & 0.96 & 4.5 & 5 & 1.0 &
79.69 & 14.06 & 6.25 &
Strongly supported \\

1 & MR-1.8 &
Should define auditor procedures &
128 & 4.17 & 0.97 & 4.0 & 5 & 1.0 &
78.12 & 14.06 & 7.81 &
Strongly supported \\

2 & MR-2.1 &
Should identify diverse stakeholder perspectives, including those affected by robustness failures based on robustness risk &
128 & 4.10 & 0.96 & 4.0 & 5 & 1.0 &
76.56 & 15.62 & 7.81 &
Strongly supported \\

2 & MR-2.2 &
Should prioritize stakeholder involvement by robustness risk &
128 & 4.00 & 1.00 & 4.0 & 5 & 2.0 &
71.88 & 19.53 & 8.59 &
Supported \\

2 & MR-2.3 &
Should include independent robustness review &
128 & 4.18 & 0.94 & 4.0 & 5 & 1.0 &
75.78 & 20.31 & 3.91 &
Strongly supported \\

2 & MR-2.4 &
Should involve experts early in robustness framing &
128 & 4.27 & 0.92 & 5.0 & 5 & 1.0 &
81.25 & 13.28 & 5.47 &
Strongly supported \\

3 & MR-3.1 &
Should scope robustness concerns and audit claims &
128 & 4.27 & 0.90 & 5.0 & 5 & 1.0 &
81.25 & 14.06 & 4.69 &
Strongly supported \\

3 & MR-3.2 &
Should identify assumptions about robustness trade-offs &
128 & 4.26 & 0.97 & 5.0 & 5 & 1.0 &
81.25 & 11.72 & 7.03 &
Strongly supported \\

4 & MR-4.1 &
Should involve relevant multidisciplinary robustness expertise &
128 & 4.32 & 0.96 & 5.0 & 5 & 1.0 &
81.25 & 14.06 & 4.69 &
Strongly supported \\

4 & MR-4.2 &
Should provide role-specific robustness training &
128 & 4.24 & 0.86 & 4.0 & 5 & 1.0 &
84.38 & 10.94 & 4.69 &
Strongly supported \\

5 & MR-5.1 &
Should ensure robustness assessment data coverage &
128 & 4.33 & 0.89 & 5.0 & 5 & 1.0 &
85.94 & 9.38 & 4.69 &
Strongly supported \\

5 & MR-5.2 &
Should document data description, provenance, and lineage &
128 & 4.23 & 0.88 & 4.0 & 5 & 1.0 &
81.25 & 13.28 & 5.47 &
Strongly supported \\

5 & MR-5.3 &
Should check data quality for robustness assessment &
128 & 4.48 & 0.89 & 5.0 & 5 & 1.0 &
87.50 & 7.81 & 4.69 &
Strongly supported \\

5 & MR-5.4 &
Should identify subgroups and data-level bias sources &
128 & 4.20 & 0.89 & 4.0 & 5 & 1.0 &
83.59 & 12.50 & 3.91 &
Strongly supported \\

5 & MR-5.5 &
Should apply robustness-focused data preprocessing mitigations &
128 & 4.02 & 0.96 & 4.0 & 4 & 1.0 &
77.34 & 14.84 & 7.81 &
Strongly supported \\

6 & MR-6.1 &
Should perform model robustness stress testing &
128 & 4.29 & 0.99 & 5.0 & 5 & 1.0 &
81.25 & 12.50 & 6.25 &
Strongly supported \\

6 & MR-6.2 &
Should specify model intended use and limits &
128 & 4.16 & 0.99 & 4.0 & 5 & 1.0 &
77.34 & 17.97 & 4.69 &
Strongly supported \\

6 & MR-6.3 &
Should support explainability for robustness assessment &
128 & 4.12 & 0.89 & 4.0 & 4 & 1.0 &
80.47 & 13.28 & 6.25 &
Strongly supported \\

6 & MR-6.4 &
Should identify model-level robustness failure sources &
128 & 4.31 & 0.99 & 5.0 & 5 & 1.0 &
84.38 & 7.03 & 8.59 &
Strongly supported \\

6 & MR-6.5 &
Should apply robustness-focused training-time mitigations &
128 & 3.94 & 0.94 & 4.0 & 4 & 2.0 &
73.44 & 21.09 & 5.47 &
Supported \\

7 & MR-7.1 &
Should identify output-level bias or failure sources and assess whether post-training mitigations effectively address them &
128 & 4.27 & 0.95 & 5.0 & 5 & 1.0 &
82.81 & 11.72 & 5.47 &
Strongly supported \\

8 & MR-8.1 &
Should integrate robustness checks into AI development workflows &
128 & 4.34 & 0.85 & 5.0 & 5 & 1.0 &
85.94 & 10.16 & 3.91 &
Strongly supported \\

8 & MR-8.2 &
Should embed robustness checks in ML/MLOps pipelines &
128 & 4.27 & 0.84 & 4.0 & 5 & 1.0 &
84.38 & 13.28 & 2.34 &
Strongly supported \\

8 & MR-8.3 &
Should ensure compatibility with AI-enabled system infrastructure &
128 & 4.27 & 0.77 & 4.0 & 4 & 1.0 &
88.28 & 7.81 & 3.91 &
Strongly supported \\

9 & MR-9.1 &
Should define human participation levels in robustness decisions &
128 & 4.18 & 0.96 & 4.0 & 5 & 1.0 &
81.25 & 12.50 & 6.25 &
Strongly supported \\

10 & MR-10.1 &
Should reassess robustness continuously or when triggered &
128 & 4.25 & 1.01 & 5.0 & 5 & 1.0 &
83.59 & 8.59 & 7.81 &
Strongly supported \\

10 & MR-10.2 &
Should track and prioritize robustness assessment progress &
128 & 4.20 & 0.88 & 4.0 & 5 & 1.0 &
79.69 & 16.41 & 3.91 &
Strongly supported \\

10 & MR-10.3 &
Should validate failures across relevant system layers &
128 & 4.27 & 0.86 & 4.0 & 5 & 1.0 &
84.38 & 12.50 & 3.12 &
Strongly supported \\

11 & MR-11.1 &
Should maintain continuous, automated robustness monitoring &
128 & 4.22 & 0.93 & 4.0 & 5 & 1.0 &
81.25 & 13.28 & 5.47 &
Strongly supported \\

11 & MR-11.2 &
Should log evidence needed to diagnose robustness failures &
128 & 4.33 & 0.85 & 5.0 & 5 & 1.0 &
84.38 & 10.94 & 4.69 &
Strongly supported \\

12 & MR-12.1 &
Should provide robustness dashboards and visualizations &
128 & 3.99 & 1.04 & 4.0 & 4 & 1.0 &
76.56 & 14.06 & 9.38 &
Strongly supported \\

12 & MR-12.2 &
Should document robustness risks, evidence, and conclusions &
128 & 4.38 & 0.87 & 5.0 & 5 & 1.0 &
86.72 & 7.03 & 6.25 &
Strongly supported \\

13 & MR-13.1 &
Should ensure robustness auditing is cost-effective, time-efficient, and scalable &
128 & 4.18 & 0.93 & 4.0 & 5 & 1.0 &
79.69 & 14.84 & 5.47 &
Strongly supported \\

14 & MR-14.1 &
Should adapt robustness assessment to the deployment domain &
128 & 4.26 & 0.85 & 4.0 & 5 & 1.0 &
85.94 & 10.16 & 3.91 &
Strongly supported \\

14 & MR-14.2 &
Should support relevant tasks, models, and modalities &
128 & 4.18 & 0.90 & 4.0 & 5 & 1.0 &
79.69 & 14.06 & 6.25 &
Strongly supported \\

15 & MR-15.1 &
Should make robustness tools, visualizations, and related outputs accessible to intended users &
128 & 4.21 & 0.99 & 4.0 & 5 & 1.0 &
81.25 & 11.72 & 7.03 &
Strongly supported \\

16 & MR-16.1 &
Should allow configuration of robustness assessment tools &
128 & 4.20 & 0.87 & 4.0 & 5 & 1.0 &
83.59 & 10.94 & 5.47 &
Strongly supported \\

\end{longtable}

\noindent\scriptsize
\textit{Note.} Agree (\%) combines \emph{Agree} and \emph{Strongly agree};
Disagree (\%) combines \emph{Disagree} and \emph{Strongly disagree}.

\end{center}

\section{Discussion} 
\label{sec:discussion} 
This section interprets the main insights arising from the literature synthesis and practitioner validation, relates them to prior work, and considers their implications for AI robustness auditing research and practice.

\subsection{A Robustness Test Is Necessary but Not Sufficient for a Robustness Audit}
\label{sec:discussion_test_audit}

The derived meta-requirements reinforce the distinction between robustness testing and auditing: tests provide
behavioural evidence, whereas an audit evaluates whether that evidence supports a defined robustness claim within specified scope, criteria, assumptions, and limitations (see Sections~\ref{sec:cat3} and~\ref{sec:cat6}). Prior work reflects this division: robustness research develops benchmarks, metrics, and evaluation methods for generating such evidence \cite{hendrycks2019benchmarking,taori2020measuring,freiesleben2023beyond}, whereas AI auditing research addresses how evidence is scoped, documented, and interpreted for assurance decisions
\cite{raji2020closing,mokander2023auditing,mokander2024auditing}. Practitioner feedback reinforces this interpretation; P123 emphasised documented evidence, limitations, and reproducibility conditions (see Section~\ref{sec:qualitative_practitioner_validation}). Thus, a successful robustness test supports an audit conclusion only when its conditions and evidence are traceable to the robustness claim being assessed.

\subsection{From Model-Level Robustness to System-Level Assurance}
\label{sec:discussion_system_assurance}

Our findings indicate that model-level robustness cannot, by itself, establish the robustness of the deployed AI-enabled system, because failures may arise from data, preprocessing, software components, output controls, or deployment workflows (see Sections~\ref{sec:cat5}-\ref{sec:cat8}). Model-level evidence must therefore be interpreted within the wider system boundary and the interactions that can affect the robustness claim. This interpretation aligns with prior work distinguishing model- from system-level assessment \cite{mokander2024auditing,nolte2025} and showing that failures can originate outside the model and propagate across components \cite{yu2026survey}. Practitioner feedback supports this view. For example, P031 noted that failures are "rarely isolated to the model" and should be assessed across the system (see Section~\ref{sec:qualitative_practitioner_validation}). Accordingly, robustness auditing should treat model-level evidence as one component of system-level assurance and bound audit conclusions to the deployed system and relevant component interactions.

\subsection{Strong Literature-Practice Convergence with Contextual Qualifications}
\label{sec:discussion_convergence}

The RQ2 findings indicate substantial convergence between the literature-derived meta-requirements and practitioner judgement, but not unconditional acceptance. Practitioner feedback largely retained the underlying concerns while refining their scope, access assumptions, feasibility, and contextual applicability (see Section~\ref{sec:qualitative_practitioner_validation}). P016's call to balance "standardization with flexibility" captures this qualification. Prior work similarly suggests that common assurance requirements can provide a shared baseline, while their operationalisation depends on system and deployment context
\cite{costanza2022audits,freiesleben2023beyond,thelisson2024conformity}. The implication is therefore not a fixed audit procedure, but a stable set of robustness auditing concerns whose implementation should remain context-sensitive. Within the studied practitioner sample, literature provides a useful foundation for \emph{what} should be addressed, while practitioner input helps determine \emph{how} those requirements should be applied.

\subsection{Robustness Is Conditional, Not Universal}
\label{sec:discussion_conditional}

Our findings indicate that robustness depends on the claim being assessed rather than being a context-free property of an AI-enabled system. The relevant system boundary, protected behaviour, operating conditions, disturbances or threats, and acceptance criteria determine what the available evidence can support (see Sections~\ref{sec:cat3} and~\ref{sec:cat14}). Practitioner feedback reinforces this interpretation: P002 called for explicit "out-of-scope" boundaries, while other respondents emphasised context-appropriate threat models, thresholds, operating conditions, and evaluation methods (see Section~\ref{sec:qualitative_practitioner_validation}). Prior work similarly treats robustness as dependent on assessment conditions and context-specific assurance criteria
\cite{freiesleben2023beyond,mokander2024auditing,thelisson2024conformity}. Accordingly, audit conclusions should state what is claimed to be robust, under which conditions, and against which criteria, rather than generalising beyond the assessed context.

\subsection{Robustness Assurance Is Time-Bounded and Requires Reassessment}
\label{sec:discussion_lifecycle}

Our findings indicate that robustness assurance remains valid only while the conditions supporting the original conclusion remain applicable. Changes to models, data, deployment conditions, threats, or other relevant factors can therefore weaken earlier evidence (see Sections~\ref{sec:cat10} and~\ref{sec:cat11}). Prior work similarly treats AI assurance as an ongoing activity in which material system changes may require renewed evaluation
\cite{mokander2024auditing,thelisson2024conformity,simbeck2024they,madaio2022assessing}.
Practitioner feedback adds an important qualification: P047 described robustness as valid only for a specific context and time, while P077 cautioned against reassessment after every minor change (see Section~\ref{sec:qualitative_practitioner_validation}). Reassessment should therefore be triggered by changes that can materially affect the robustness claim, with its scope proportionate to their significance and risk. Robustness assurance is thus better treated as a maintained conclusion than as a one-time judgement.

\subsection{An Auditable AI-Enabled System Requires an Audit-Ready Organisation}
\label{sec:discussion_audit_readiness}

Our findings indicate that technical auditability alone is insufficient for credible robustness auditing. Governance, engagement, competence, and human-oversight requirements show that robustness evidence must be interpreted and acted on by people with appropriate expertise, independence, authority, and defined responsibilities (see Sections~\ref{sec:cat1}, \ref{sec:cat2}, \ref{sec:cat4}, and~\ref{sec:cat9}). Prior work similarly frames AI auditing as an organisational and interdisciplinary activity rather than a purely technical assessment
\cite{raji2020closing,costanza2022audits,chander2025toward,li2025artificial}.
Practitioner feedback reinforces this interpretation: P003 emphasised early expert involvement and independence, while P006 highlighted documented stakeholder input and resolution of conflicting viewpoints (see Section~\ref{sec:qualitative_practitioner_validation}). Accordingly, audit readiness should cover both the AI-enabled system and the organisation by defining required expertise, independence, responsibilities, decision authority, and procedures for handling disagreements and follow-up actions.

\subsection{External Components Extend the Scope of Robustness Audits}
\label{sec:discussion_external_dependencies}

Our findings indicate that robustness assurance may depend on components outside the audited organisation's direct control. External models, APIs, datasets, libraries, plugins, and cloud services can affect the robustness claim while limiting auditor visibility and access to evidence. Practitioner feedback exposed this gap: P079 highlighted "third-party and supply chain risks," while P102 identified foundation models and third-party APIs as requiring explicit audit attention (see Section~\ref{sec:qualitative_practitioner_validation}). Prior work similarly shows that assurance evidence may be distributed across providers, deployers, and auditors
\cite{mokander2024auditing,thelisson2024conformity,laux2024trustworthy}.  MR-8.4 therefore extends the audit boundary to relevant external dependencies by requiring the audit to identify them, assess how their failure or change could affect the robustness claim, and state how limited access constrains the robustness conclusion \cite{dev2025building,yu2026survey,martinez2022software,blood2023reliability,casper2024black,mokander2024auditing}. Relevant records for identifying and tracing these external dependencies may include SBOMs and AIBOMs \cite{nocera2025we,radanliev2026operationalising}. Because MR-8.4 emerged after the original requirements were rated, its practical value requires further empirical evaluation.

\subsection{Implications for Practice}
\label{sec:discussion_practice}

The meta-requirements are intended to support the completeness and design of AI robustness audits by specifying what should be defined, tested, documented, assigned, monitored, and reassessed. They may also complement instruments such as the EU AI Act and NIST AI RMF ~\cite{thelisson2024conformity,li2025artificial}, although no systematic mapping to their provisions was performed.

\subsubsection{For AI Robustness Auditors}
\label{sec:implications_robustness_auditors}

Our findings imply that robustness audits should begin with a clearly defined robustness claim rather than an available test. Auditors should establish the system boundary, operating conditions, relevant threats or disturbances, access conditions, metrics, and decision criteria before selecting evaluation methods \cite{raji2020closing,mokander2024auditing}. For example, these requirements may translate into distribution-shift tests and reassessment triggers for an image classifier, or jailbreak conditions, tool interactions, and dependency evidence for an LLM-based system. Accordingly, audit findings should remain traceable to the supporting evidence and evaluated conditions, while limitations in evidence or access should be stated explicitly (see Sections~\ref{sec:discussion_test_audit} and \ref{sec:discussion_conditional}).

\subsubsection{For Providers and Deployers of AI-Enabled Systems}
\label{sec:implications_providers_deployers}

Our findings indicate that providers and deployers should maintain audit-ready evidence throughout the AI-enabled system lifecycle. This includes preserving relevant system versions, data provenance, configurations, evaluation settings, monitoring records, incidents, external dependencies, and robustness decisions \cite{raji2020closing,simbeck2024they,li2025artificial}. Accordingly, changes that may affect a robustness claim should be traceable to the reassessment undertaken and any resulting follow-up actions (see Sections~\ref{sec:discussion_system_assurance} and \ref{sec:discussion_lifecycle}).

\subsubsection{For Builders of Robustness Auditing Tools}
\label{sec:implications_tool_builders}

Our findings indicate that robustness testing tools are insufficient on their own for AI robustness auditing. Existing toolkits can generate adversarial examples, run perturbation tests, measure robustness, or identify vulnerabilities \cite{agarwal2025advancing}. However, tools that support robustness auditing should also link the robustness claim and scope to the test configuration, system versions, supporting evidence, findings, mitigation status, retesting, and reassessment \cite{raji2020closing}. Accordingly, different user views should remain linked to the same audit evidence record to support consistent, traceable reporting.

\subsubsection{For Standards and Certification Bodies}
\label{sec:implications_standards_certification}

Our findings suggest that standards and certification schemes should define common requirements for substantiating robustness claims rather than mandate a single test, metric, or threshold. These requirements should cover terminology, claim definition, minimum evidence, access disclosure, auditor competence, reporting, reassessment, and closure of audit findings \cite{thelisson2024conformity,nolte2025}. However, assessment methods and thresholds should remain adaptable to the domain, system type, risk level, and deployment context (see Section~\ref{sec:discussion_conditional}). Accordingly, standards should require such choices and deviations to be justified and should state how restricted access or external dependencies limit the assurance that can be claimed.

\subsection{Implications for Research on Robustness Auditing}
\label{sec:discussion_research_implications}

The findings also identify several directions for robustness auditing research. In particular, practitioner validation establishes the relevance and interpretation of the meta-requirements, but further work is needed to determine how they should be operationalised and whether their use improves audit practice.

\subsubsection{Operationalising the Meta-Requirements through a Robustness Auditing Artefact}
\label{sec:research_operationalising_artefact}

This study identifies and validates meta-requirements for AI robustness auditing but does not specify how to implement them. Future design-science research should therefore translate them into a concrete audit artefact that defines procedures, links robustness claims to evidence, specifies access and decision rules, assigns responsibilities, and supports reassessment. Building on structured AI auditing processes \cite{raji2020closing,mokander2024auditing}, such an artefact should be evaluated for usability, evidence completeness, traceability, and integration with existing development, MLOps, governance, and assurance processes. This would establish how the meta-requirements can be applied consistently while allowing adaptation to different AI-enabled systems and audit contexts.

\subsubsection{Evaluating the Meta-Requirements in Real Audit Engagements}
\label{sec:research_real_audit_evaluation}

Practitioner validation establishes perceived relevance, but it does not show that applying the meta-requirements improves the quality of an actual robustness audit. Future case studies and action-research engagements should therefore apply the meta-requirements to AI-enabled systems that vary in domain, architecture, risk, lifecycle stage, and audit-access level. Such studies could examine audit coverage, evidence completeness, traceability, effort, finding quality, follow-up actions, and decision usefulness. This would extend prior work calling for structured AI auditing processes to be evaluated within organisational practice \cite{raji2020closing,mokander2024auditing}. The objective is to determine which meta-requirements are practically effective, which remain difficult to operationalise, and where contextual adaptation is necessary without weakening the evidential basis of the audit.

\subsubsection{Assessing Inter-Auditor Consistency and Repeatability}
\label{sec:research_inter_auditor_consistency}

A further open question is whether the meta-requirements help different auditors reach sufficiently consistent and explainable conclusions. The present validation examined practitioner support for the meta-requirements, not their consistency of application. Future studies could therefore ask multiple auditors to assess the same AI-enabled system and compare the scopes they define, the evidence they request, the tests they select, the findings they identify, and the conclusions they reach. Where feasible, comparison with audits conducted without guidance from the meta-requirements could test whether their use improves traceability, transparency, and repeatability. The goal should not be identical audit reports, since professional judgement may legitimately differ across contexts, but rather to determine whether the meta-requirements makes such differences explicit and defensible through documented scope, assumptions, evidence, access limitations, and decision criteria.

\section{Threats to Validity}
\label{sec:threats_validity}

We considered potential threats to external, construct, internal, and conclusion validity~\cite{wohlin2012experimentation} arising from the literature synthesis and practitioner validation.

\subsection{External Validity}
\label{sec:external_validity}

External validity concerns the extent to which the findings generalise beyond the reviewed literature and practitioner sample~\cite{wohlin2012experimentation}. Relevant studies may have been missed because AI robustness auditing spans multiple research communities and terminology, and relevant evidence may exist outside academic literature.
In particular, the practitioner-oriented and interview or survey search terms may have underrepresented technically relevant robustness studies that did not use such terminology. We mitigated this through multiple database searches, pilot-refined search strings, supplementary Google Scholar screening, and explicit eligibility criteria (see Section~\ref{sec:literature_search}). Nevertheless, the 74 retained studies represent a systematically selected academic evidence base rather than an exhaustive account; proprietary, unpublished, non-English, and other non-academic practices may be underrepresented. We recruited participants for the main survey through GitHub and LinkedIn using purposive non-probability sampling (see Section~\ref{sec:participant_recruitment}), introducing possible self-selection and coverage bias. Although we targeted participants with relevant AI/ML, software engineering, robustness, governance, monitoring, and auditing backgrounds and collected profile information to characterise the sample, it is not statistically representative of the wider practitioner population. The findings may therefore transfer differently across stakeholder groups, domains, system types, and organisational settings.

\subsection{Construct Validity}
\label{sec:construct_validity}

Construct validity concerns whether the study adequately captures the concepts it investigates~\cite{wohlin2012experimentation}. Concepts such as robustness, auditing, monitoring, and assurance may be interpreted differently across research and professional communities, and the technology-agnostic formulation of the meta-requirements may have led respondents to interpret their operational meaning differently. Moreover, a five-point Likert rating captures agreement but does not establish shared interpretation, implementation feasibility, or practical effectiveness. We mitigated these risks by grounding the meta-requirements in source-traceable literature evidence (see Section~\ref{sec:mr_extraction}), organising them into explicit audit categories, and piloting the survey for terminology, wording, presentation, and completion effort (see Section~\ref{sec:practitioner_validation}). Likert ratings were complemented by open-text feedback on ambiguity, overlap, feasibility, and missing issues. Nevertheless, some interpretive variation remains across heterogeneous professional and system contexts.

\subsection{Internal Validity}
\label{sec:internal_validity}

Internal validity concerns potential bias introduced by the research process~\cite{wohlin2012experimentation}. Study screening and the derivation of meta-requirements required researcher judgement, particularly for borderline studies, overlapping concepts, and abstraction decisions. We mitigated this through explicit eligibility criteria, independent screening by two researchers, Cohen’s $\kappa$ assessment before reconciliation ($\kappa=0.882$ for title/abstract and $\kappa=0.741$ for full text), consensus resolution of screening disagreements, source-to-requirement traceability, and review by a second researcher of extracted patterns, category assignments, and meta-requirement formulations, with remaining ambiguities resolved through discussion (see Sections~\ref{sec:literature_search} and~\ref{sec:mr_extraction}). Nevertheless, other researchers may make different decisions for some studies or formulate particular meta-requirements differently. The practitioner phase may also be affected by response and interpretation bias. Rating 43 meta-requirements could cause fatigue or satisficing, while the desirable nature of robustness, governance, and auditability may encourage favourable responses. Pilot testing addressed clarity and completion effort, and open-text feedback on qualifications, implementation concerns, and challenges complemented the agreement ratings. Qualitative coding still involved researcher interpretation; despite structured coding and traceable coding and synthesis decisions (see Section~\ref{sec:qualitative_analysis}), some comments may reasonably be interpreted or mapped differently.

\subsection{Conclusion Validity}
\label{sec:conclusion_validity}

Conclusion validity concerns whether the conclusions are supported by the analysed evidence~\cite{wohlin2012experimentation}. High practitioner agreement supports the literature-derived meta-requirements but does not establish that applying the meta-requirements improves audit quality, robustness outcomes, or inter-auditor consistency. Likewise, open-text comments capture how respondents interpreted and refined the meta-requirements, rather than prevalence in the wider practitioner population. We therefore analysed the ratings descriptively and integrated them with qualitative evidence rather than treating quantitative agreement as sufficient validation. We compared candidate additions against the existing meta-requirements to avoid isolated or overlapping requirements. This synthesis produced one new requirement, MR-8.4, which was not quantitatively rated because it emerged after the original 43 meta-requirements had been evaluated. Thus, the study provides content-oriented practitioner validation rather than evidence of operational effectiveness; further evaluation through real audit engagements and inter-auditor studies is required.

\section{Conclusion and Future Work}
\label{sec:conclusion}

AI robustness auditing knowledge is distributed across multiple research domains, including AI, software engineering, and auditing. This study addressed this gap by deriving literature-grounded meta-requirements for AI robustness auditing and evaluating them with practitioners. The systematic literature review retained 74 studies and produced 43 literature-derived meta-requirements across 16 categories. The findings show that robustness auditing extends beyond model-level testing to scoping robustness claims, assessing available evidence and system components, documenting limitations, and continuing monitoring and reassessment. The practitioner survey with 128 participants showed broad support, with 81.20\% of ratings indicating agreement or strong agreement and 41 of the 43 literature-derived requirements classified as strongly supported. Practitioner feedback nevertheless refined requirement definitions, scope, implementation conditions, stakeholder responsibilities, workflow integration, monitoring, and reporting. Integrating both evidence sources resulted in a final practitioner-informed set of 44 meta-requirements across 16 categories, including one additional requirement addressing external components and supply chain dependencies. The final set of meta-requirements organises the identified auditing concerns into 44 meta-requirements covering areas such as audit scope and responsibilities, data and model assessment, evidence, workflow integration, monitoring, reporting, and reassessment. Future work will operationalise the meta-requirements into a decision model for selecting appropriate requirements, evidence, tests, metrics, and assessment conditions for a given AI-enabled system and audit context, and subsequently integrate this model into a design support system for practical AI robustness auditing.

\section*{Declaration of Competing Interest}

The authors declare that they have no known competing financial interests or personal relationships that could have appeared to influence the work reported in this paper. This research received funding from Deloitte, as disclosed in the Acknowledgments. Deloitte had no role in the study design, data collection, analysis, interpretation of results, or decision to submit the manuscript for publication.

\section*{Acknowledgments}

This research has been supported by the Safeguard project (grant No.\ 20506020). It has also received funding from Deloitte. The authors thank all practitioners who participated in the validation survey and provided feedback on the proposed AI robustness auditing meta-requirements. We also thank colleagues and professional networks who helped distribute the pilot survey, as well as those who provided constructive comments on earlier versions of the manuscript.

\section*{Ethics Approval}

The practitioner survey received ethical clearance from the Research Ethics and Data Management Committee of the Tilburg School of Humanities and Digital Sciences, Tilburg University (REDC 2026.04). We obtained informed consent from all survey participants before participation.


\appendix

\section*{Data Availability Statement}

For transparency and reproducibility, we provide a replication package\footnote{\url{https://doi.org/10.5281/zenodo.23267777}} containing the review protocol, search strings, included-study and quality-assessment records, extraction and coding records, meta-requirement tables, survey instrument, practitioner-validation data, and analysis outputs.



\bibliographystyle{ACM-Reference-Format}
\bibliography{references}

\end{document}